\documentclass[letterpaper, 10 pt, journal, twoside]{IEEEtran}
\usepackage{amsmath}
\usepackage{amssymb}
\usepackage{bm}
\usepackage{graphicx}
\usepackage{subcaption}
\usepackage[skip=0pt,font=small,labelfont=bf]{caption}
\usepackage{wrapfig}
\usepackage{adjustbox}
\usepackage{pgfplots}
\usetikzlibrary{patterns}
\usepackage{color,soul}
\usepackage{amssymb}
\usepackage{url}
\usepackage{breqn}
\usepackage{multirow}
\usepackage[top=72pt, left=54pt, right=54pt, bottom=54pt]{geometry}
\usepackage{cite} % Group multiple citations into brackets
\usepackage[ruled,linesnumbered]{algorithm2e}
\usepackage[noend]{algpseudocode}

\usepackage{fixme}
\fxsetup{status=draft, theme=color}
\definecolor{fxtarget}{rgb}{0.8000,0.0000,0.0000}
\definecolor{fxnote}{rgb}{0.0000,0.0000,0.8000}

\begin{document}

\title{\LARGE \bf
	Excavation Learning for Rigid Objects in Clutter
}

% \author{Qingkai Lu$^{1}$ and Liangjun Zhang$^{1}$%

% \thanks{$^{1}$Qingkai Lu and Liangjun Zhang are
%   with the Robotics and Auto-Driving Lab, Baidu Research, Sunnyvale, CA USA. 
%  {\tt\footnotesize qingkailu@baidu.com;
%  	liangjunzhang@baidu.com}.}%
% }

\author{Qingkai Lu$^{1}$ and Liangjun Zhang$^{1}$%
\thanks{Manuscript received: February, 24, 2021; Revised May, 22, 2021; Accepted June, 23, 2021.}%Use only for final RAL version
\thanks{This paper was recommended for publication by Editor M. Vincze upon evaluation of the Associate Editor and Reviewers' comments.
This work was supported by Baidu Research USA.} %Use only for final RAL version
\thanks{$^{1}$Qingkai Lu and Liangjun Zhang are with the Robotics and Auto-Driving Lab, Baidu Research, Sunnyvale, CA USA. 
        {\tt\footnotesize qingkailu@baidu.com; 
        liangjunzhang@baidu.com}}%
\thanks{Digital Object Identifier (DOI): see top of this page.}
}

\markboth{IEEE Robotics and Automation Letters. Preprint Version. Accepted June, 2021.}
{Lu \MakeLowercase{\textit{et al.}}: Excavation Learning for Rigid Objects in Clutter}

\maketitle
%\thispagestyle{empty}
%\pagestyle{empty}

% As a general rule, do not put math, special symbols or citations
% in the abstract or keywords.
\begin{abstract}
Autonomous excavation for hard or compact materials, especially irregular rigid objects, is challenging due to high variance of geometric and physical properties of objects, and large resistive force during excavation. In this paper, we propose a novel learning-based excavation planning method for rigid objects in clutter. Our method consists of a convolutional neural network to predict the excavation success and a sampling-based optimization method for planning high-quality excavation trajectories leveraging the learned prediction model. To reduce the sim2real gap for excavation learning, we propose a voxel-based representation of the excavation scene. We perform excavation experiments in both simulation and real world to evaluate the learning-based excavation planners. We further compare with two heuristic baseline excavation planners and one data-driven scene-independent planner. The experimental results show that our method can plan high-quality excavations for rigid objects in clutter and outperforms the baseline methods by large margins. As far as we know, our work presents the first learning-based excavation planner for cluttered and irregular rigid objects. 
\end{abstract}

\begin{IEEEkeywords}
Deep learning in grasping and manipulation, perception for grasping and manipulation, manipulation planning
\end{IEEEkeywords}

% Note that keywords are not normally used for peerreview papers.

%\begin{IEEEkeywords}
%\end{IEEEkeywords}

%\vspace{-5pt}
\section{Introduction}
\label{sec:intro}
\IEEEPARstart{E}{xcavators} are widely used in various applications, including construction, material loading, and mining. Excavators need to operate in extreme environments or weather conditions, which are challenging for human operators.
Operating excavators requires special and costly training to ensure safe operations of equipment~\cite{wang2007design}. Moreover, occupational machine-related fatalities and injuries occur each year~\cite{Marsh2015}.
Automating the excavator operation has been an active area of research because of its potential to increase safety, reduce cost and improve the work efficiency~\cite{sing1995synthesis,stentz1999robotic,jud2017planning}. 
%there is a lack of skilled operators for operating excavators.

In terms of developing autonomous excavator systems, there have been many efforts that focus on particular aspects~\cite{hemami2009overview, dadhich2016key}, including perception~\cite{mascaro2020towards}, planning~\cite{Seo2015TaskPD,yang2020Opt}, control~\cite{jud2017planning}, teleoperation~\cite{tripicchio2017stereo, tanzini2013novel}, and system integration and applications~\cite{stentz1999robotic}. Despite these advances, autonomous excavation for hard or compact materials, especially irregular rigid objects, remains challenging and relatively few works have looked at this problem~\cite{fernando2019iterative, dobson2017admittance}. 

Rock excavations are typical scenarios in mining job sites~\cite{Sean03}. As compared to granular material, rocks are rigid and often in clutter. It is more challenging, more time consuming, and much more expensive to excavate~\cite{Sean03} rocks. Excavation of rocks can result in large resistive forces to the bucket~\cite{marshall2008toward}. Furthermore, unlike granular materials composited by uniform particles, rigid objects often have high variance of geometrical shapes (e.g., concave and convex), appearances, and physics properties (e.g., mass), which largely increases the challenges for robotic perception and manipulation.

In this paper, we focus on excavation learning and planning for irregular rigid objects in clutter. We employ deep learning to tackle the challenges of excavation for rigid objects in clutter. Given the visual representation of the excavation scene, our goal is to plan a high-quality trajectory to excavate objects with large total volume per excavation.
We first propose novel RGBD and voxel-based convolutional neural network (CNN) models to predict the excavation success, which we train by collecting a large set of training excavation samples in simulation. We then formulate the excavation planning as an optimization problem leveraging the learned prediction models. We perform excavation experiments in both simulation and real world to evaluate our learning-based excavation methods. Our excavation experiments in simulation and real world show that our learning-based planners are able to generate excavations with high success rates. The experimental results also demonstrate the advantages of the learning-based excavation planners over two heuristic planners.

\vspace{-2pt}
In summary, the main contribution of this work includes:
\begin{enumerate}
\item We propose two CNN models for success prediction of a new task, excavation for rigid objects in clutter, and solve the excavation planning as an optimization problem leveraging the learned models. 
\item Our excavation experiments in simulation and real world show that our learning-based planner is able to generate excavation trajectories with a high success rate.
\item We represent the excavation trajectories in task space, which allows the transfer of the learned excavation prediction models across different hardware platforms. 
\item We demonstrate the voxel-grid representation of the excavation scene reduces the sim2real gap for excavation learning, compared with the RGBD image representation.
\item We collected and released an excavation dataset for cluttered rigid objects. %\textbf{TODO: Provide the data link.}
\end{enumerate}
\vspace{-2pt}

We summarize the related work in Section~\ref{sec:related_work}. In Section~\ref{sec:problem_define}, we define the excavation planning problem for cluttered rigid objects. We follow this in Section~\ref{sec:excavation_learning} with an overview of our approach to excavation learning and planning. We then give a thorough account of our simulated and real-robot experiments in Section~\ref{sec:experiments}. We conclude with a brief discussion in Section~\ref{sec:discussion}.
In the Appendix (i.e., Section~\ref{sec:appendix}), we present the excavation data collection, model training, offline validation, further results analysis, and ablation study.

\vspace{-10pt}
\section{Related Work}
\label{sec:related_work}
In this section, we summarize the literature of autonomous excavators, manipulation learning, and voxel-based planning.

\vspace{-7pt}
\subsection{Autonomous Excavators}
Prior work on developing autonomous excavators mainly focuses on soil excavation and granular material handling. In the seminal work~\cite{stentz1999robotic}, a prototype system for autonomous material loading to dump trucks is proposed. Recently, a system for autonomous trenching~\cite{jud2019autonomous} is presented and validated on a real excavator. 
Yang and colleagues~\cite{yang2020Opt} propose a trajectory optimization method for for granular material excavation.
Recently, various prototypes and experiments have been carried out on the task planning for large-scale excavation tasks, e.g., soil pile removals \cite{Seo2015TaskPD}. 
A novel real-time panoramic telepresence system for construction machines is presented in~\cite{tripicchio2017stereo}.
Tanzini and others discuss a novel approach for interactive
operation of working machines in~\cite{tanzini2013novel}.
A reinforcement learning approach is proposed for automated arm control of a hydraulic excavator in~\cite{tanzini2013novel}.

% maeda2015combined

Different control approaches have been proposed for excavation automation. 
Maeda et al. propose a new control structure with explicit disturbance compensation for soil excavation in~\cite{maeda2015combined}.
In \cite{jud2017planning}, a force control method is presented and the resulting bucket motions can be adaptive to different terrain. In~\cite{chang2002straight}, a straight-line motion tracking control scheme is proposed for hydraulic excavator system. In~\cite{sotiropoulos2019model}, a model-free extremum-seeking approach using power maximization is presented. 

%{fernando2019iterative, dobson2017admittance}
There are relatively few work related to rigid objects excavation~\cite{fernando2019iterative, dobson2017admittance, sotiropoulos2020autonomous}.
Fernando and others develop an iterative learning-based admittance control algorithm for autonomous excavation in fragmented rock using robotic wheel loaders.
An admittance-based Autonomous Loading Controller for fragmented rock excavation is discussed in~\cite{dobson2017admittance}. Compared with the low-level excavation control work in~\cite{fernando2019iterative, dobson2017admittance}, our work focuses on learning-based excavation trajectory planning that considers the visual scene representation of cluttered rigid objects. 
In~\cite{sotiropoulos2020autonomous}, Sotiropoulos and Asada integrate a Gaussian process rock motion model and an Unscented Kalman Filter for rock excavation.
% The method is validated using UR5 robot mounted with a 3D printed bucket. 
However, they only focus on excavation of a single rock in isolation and use the OptiTrack motion capture system to track the motion of the rock. In comparison, we focus on excavation for rigid objects in clutter using a RGBD camera.
%, which are more general in terms of excavation scenario setup.

% In \cite{jud2017planning} a terrain adaptive navigation was developed for walking  excavators. 

% \cite{hutter2015towards}
% \cite{hutter2016force}
% \cite{hutter2016ibex}

\vspace{-2pt}
\subsection{Deep Learning for Manipulation} 
% \textbf{TODO: the advantage and benefits of robot manipulation learning; the motivation of excavation learning for rigid objects in clutter.}
% Learning-based approaches to grasping [1–8] have become
% a popular alternative to geometric [9–12] and model-based
% planning [13, 14] over the past decade. In particular grasp
% learning has shown to generalize well to previously unseen
% objects where only partial-view visual information is available.
% More recently, researchers have looked to capitalize on the
% success of deep neural networks to improve grasp learning.

In recent years, researchers have looked to capitalize on the success of deep learning to improve robotic manipulation, including non-prehensile manipulation~\cite{finn2017deep},
%xu2020learning
grasping~\cite{levine2018learning, lu2017grasp, lenz2015deep}, and granular material scooping~\cite{schenck2017learning,clarke2018learning}. 
For example, deep learning has been shown to generalize well to previously unseen objects where only partial-view visual information is available for grasping~\cite{lu2020multi}.
In~\cite{halbach2019neural}, Halbach and others train an end-to-end Neural Network controller for automated pile loading of granular media using human demonstration data. 
In \cite{sandzimier2020data}, a statistical model is learned to predict the behavior of soil excavation, which is used for controlling the amount of excavated soil. In our work, we apply deep learning to tackle the perception and manipulation challenges of excavation for cluttered rigid objects and generate high-quality excavations. 
% In~\cite{clarke2018learning}, Clarke et al. propose learning frameworks for estimating amounts and flows of granular material from audio data during robotic pouring and scooping tasks. 

%  The excavation process is modelled as a heteroscedastic Gaussian process (GP). 

% Others: \cite{zeng2019tossingbot}
Various planning approaches have been developed to leverage deep neural network predictive models. In~\cite{lenz2015deep}, Lenz and colleagues proposes cascaded deep networks to efficiently evaluate a large number of candidate grasps. 
In~\cite{schenck2017learning}, a highly-tailored CNN model is developed to learn the dynamics of the granular material scooping task and the cross entropy method (CEM) leveraging the learned prediction model is used for scoop planning.
In~\cite{lu2017grasp}, the grasping planning is formulated and solved as a gradient-based optimization over the grasp configuration leveraging the grasp predication network.
%which leverages the efficient computation of gradients in neural networks. 
In this paper, we model excavation planning as an optimization problem which maximizes the probability of excavation success predicted by our excavation prediction network and solve the optimization using CEM.

% Manipulation + motion planning: \cite{wang2019manipulation}

% General excavation.  Traditional excavation methods.
% Excavation for rigid objects.
% Excavation learning. 
% Non-prehensile manipulation, such as pushing. 

% Deep learning is applied to learn robotics scooping of granular material in~\cite{schenck2017learning, clarke2018learning}.
% In~\cite{clarke2018learning}, Clarke and colleagues propose learning frameworks for estimating amounts and flows of granular material from audio data during robotic pouring and scooping tasks. %They evaluate multiple deep and shallow learning frameworks on a shaking and pouring dataset across different granular materials. Their results indicate that audio is an informative sensor modality to estimate flow and amounts of pouring. 

\vspace{-8pt}
\subsection{Voxel-based Planning}
In~\cite{lu2020multi}, a voxel-based object representation and two 3D CNNs are presented for multi-fingered grasp learning and planning.
Jetchev and Toussaint model environments with voxel-grids and present a novel way for faster movement planning in such environments by predicting good path
initializations~\cite{jetchev2010trajectory}.
To overcome the sim2real issue, we propose a 3D voxel-grid representation of the excavation scene.

%\vspace{-5pt}
\section{Problem Definition}
\label{sec:problem_define}
In this section, we define the excavation task for rigid objects and the excavation trajectory representation. 

\vspace{-2pt}
\subsection{Task Overview}
\label{sec:task_overview}
This work focuses on rigid objects excavation in clutter. Given the visual representation (i.e., the RGBD image or voxel-grid in this work) $Z$ of the current excavation scene, our goal is to plan a trajectory $T$ that excavates rigid objects (e.g., stones or wood blocks) with the maximum total volume $V$. An excavation instance/sample is defined to be the pair of the scene visual representation and the excavation trajectory $(Z, T)$. We focus on maximizing the excavated objects volume of the current excavation greedily without considering the future excavations. 
% The excavation task can potentially be further extended as maximizing the accumulated excavated objects volume with a fixed number of excavations or emptying the target region of objects using  the least number of excavations. 

\begin{figure}
    \centering
    \begin{subfigure}[b]{0.23\textwidth}
        \centering
        \includegraphics[width=\textwidth]{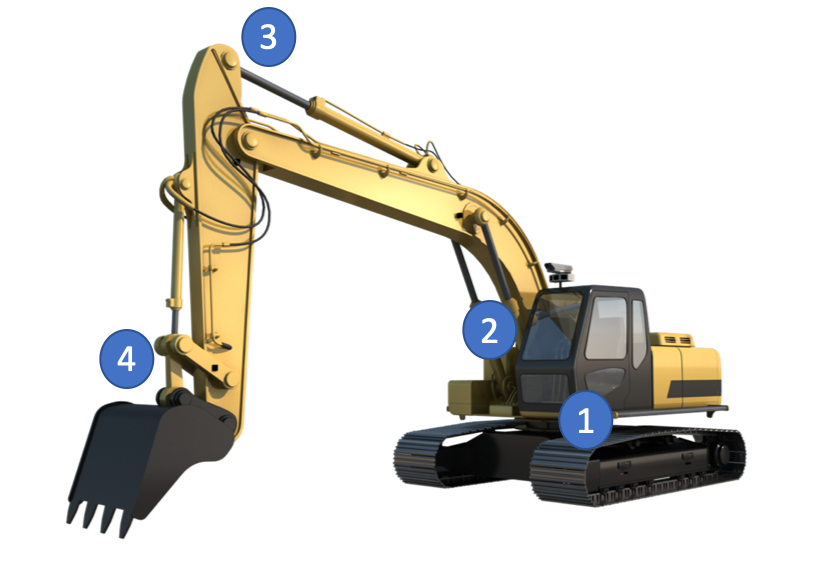}
        \caption{An excavator.}
        \label{fig:real_exv_kin}
    \end{subfigure}
    \hfill
    \begin{subfigure}[b]{0.23\textwidth}   
        \centering 
        \includegraphics[width=\textwidth]{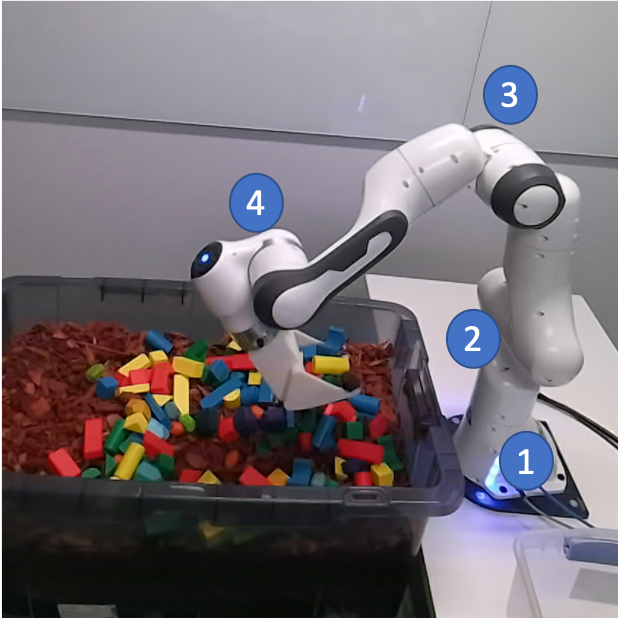}
        \caption{Our Franka ``excavator".}  
        \label{fig:franka_exv_kin}
    \end{subfigure}
    \caption{The mechanical structure of an excavator and our Franka arm ``excavator" with a 3D printed bucket are shown in this figure. The first to forth joint of the excavator and Franka are the base swinging, boom, stick, and bucket joint respectively.} 
    \label{fig:exv_kin}
\end{figure}

In this work, we emulate a standard 4 DOF excavator model using a Franka Panda 7 DOF robot arm mounted with a 3D printed excavation bucket. Figure~\ref{fig:franka_exv_kin} shows our excavation task and scene setup, where rigid wood blocks are put in an excavation tray and the task is to excavate these blocks and dump to a dumping tray after each excavation.

% is mounted to the Franka arm as the excavation end-effector. We put wooden and plastic rigid toy blocks with different geometries and colors in an excavation tray for excavation experiments in real world. The Azure RGBD camera is used to get the RGB and depth images of the excavation scene. The excavation setup in real world is shown in Figure~\ref{fig:franka_exv_kin}.

% The robot dumps the excavated objects into a dumping tray after each excavation.

% \begin{figure}
%     \centering
%     \includegraphics[width=0.4\textwidth]{figures/excavator_ab.png}
%     \caption{We show the kinematics of an excavator (a) and our Franka arm ``excavator'' (b) with a 3D printed bucket in this figure. The first to forth joint of the excavator and Franka are the base swinging, boom, stick, and bucket joint respectively.}
%     \label{fig:exv_kin}
% \end{figure}

% The first to forth joint of the excavator and Franka are the shoulder swinging joint, the shoulder lifting joint, the elbow lifting joint, and the wrist lifting joint respectively.

\vspace{-10pt}

%\vspace{-5pt}
\subsection{Excavation Trajectory Representation in Task Space}
\label{sec:task_traj}
\begin{figure}[h]
    \centering
    \includegraphics[width=0.45\textwidth]{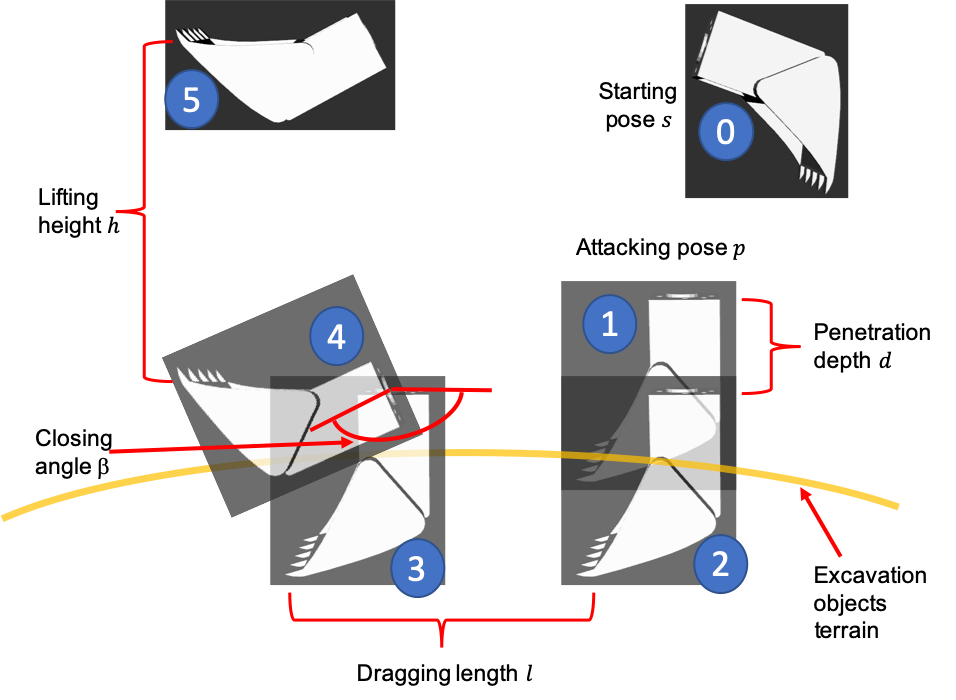}
    \caption{This figure visualizes our excavation trajectory representation in the task space. The numbers in blue circles represent the sequence of our five excavation phases. The bucket pose with ID $0$ shows the starting pose. The bucket poses with ID from $1$ to $5$ represents the attacking, penetration, dragging, closing, and lifting phase respectively.}
    \label{fig:exv_task_traj}
\end{figure}

% An excavator arm usually has 4 DoF as can be seen from Figure~\ref{fig:exv_kin}, including the shoulder panning/swinging joint, the shoulder lifting joint, the elbow lifting joint, and the wrist lifting joint.
As shown in Figure~\ref{fig:exv_kin}, an excavator arm usually has 4 DoF, including the base swinging, boom, stick, and bucket joint.
The $4$D excavation pose of the bucket in task (Cartesian) space is composed of the $3$D excavation position $(x, y, z)$ and the $1$D excavation angle $\alpha$. The excavation angle determines the bucket orientation, which also equals to the sum of the joint angles of the last three excavation joints. We define the excavation angle to be zero degree when the bucket orientation is horizontal and points away from the robot. The excavation angle is $-90$ degree when the bucket orientation is vertical and points down. One example of the excavation angle visualization can be seen from the closing excavation angle $\beta$ in Figure~\ref{fig:exv_task_traj}. 

In general, an excavation trajectory $T$ can be divided into multiple phases~\cite{sing1995synthesis}. Figure~\ref{fig:exv_task_traj} illustrates one scheme, where the trajectory is divided into 5 phases: attacking, penetration, dragging, closing, and lifting. In the attacking phase, the excavator arm moves the bucket from the starting pose to its $4$D target attacking pose $\mathbf{p}=(x, y, z, \alpha)$. In the penetration phase, the bucket penetrates into objects with a specified depth $d$ along the gravity direction. 
% \liangjun{ Here you assume the penetration direction is the vertical and the dragging direction is the horizontal. This is a simplified way. Also, for penetration, it is often desired that the bucket heading is perpendicular to the soil surface to minimize the residence force.   } 
Then the bucket drags horizontally towards the excavator base in the excavation plane for a given length $l$. Dragging allows the excavator arm to push and accumulate more objects into the bucket along the way. 
% In the closing phase, the excavator arm decreases the angle between the bucket and the horizontal plane to manipulate objects into the bucket and close the bucket. The excavator arm closes the bucket by decreasing the wrist joint angle to $\beta$ degree. 
In the closing phase, the excavator arm decreases the angle between the bucket and the horizontal plane to $\beta$ degree in order to close the bucket and manipulate objects into the bucket. 
%The excavator arm closes the bucket by decreasing the wrist joint angle to $\beta$ degree.
Finally, the excavator arm lifts the bucket to a certain height $h$. 

%\vspace{-7pt}
\vspace{-3pt}

We assume the attacking point is always on the surface of the objects clutter. Given the $(x, y)$ coordinate of the attacking pose $p$, its $z$ coordinate value on the objects clutter surface is computed as the height of the grid/height map of the objects clutter at $(x, y)$. More details and examples of the grid map can be found in Section~\ref{sec:data_collection} of the Appendix. We fix the lifting height $h$ to lift the bucket to the height of the robot base. Therefore, we can represent a task space excavation trajectory $T$ using $6$ parameters $T=(x, y, \alpha, d, l, \beta)$. The points of attack $(x, y)$ are learned and planned in the objects tray frame. The object tray frame definition for simulation and real world is explained in more detail in Section~\ref{sec:exp_setup_sim} and~\ref{sec:exv_scene_setup_real} of the Appendix respectively. 

Having the $6$D task trajectory parameters, we interpolate the excavation trajectory and generate its corresponding joint space trajectory by applying inverse kinematics (IK) of the excavator arm~\cite{yang2019compact}. Then the interpolated joint trajectory waypoints are sent to a position controller in both simulation and real world.
Though the robotic arm is used for excavation in this paper, the task trajectory representation and excavator IK can also be translated directly to hydraulically actuated excavator arms.

%Then we interpolate trajectory waypoints for the attacking and closing phase with an interpolation step size of $5$ degree in the joint space. For the penetration, dragging, and lifting phase, we generate trajectory waypoints with an interpolation step size of $2cm$ in the task space. Then we compute the joint configurations of the trajectory waypoints of these $3$ phases using IK. Finally, the interpolated excavation joint space trajectory is sent to a position controller in both simulation and real world.

% For the penetration, dragging, and lifting phases, we generate trajectory waypoints with an interpolation of the end-effector and computing the joint configurations of the trajectory waypoints of these $3$ phases using IK. The interpolated excavation trajectory is sent to a position controller in both simulation and real world.
% with an interpolation step size of $5$ degree in the joint space.
% For the penetration, dragging, and lifting phase, we generate trajectory waypoints with an interpolation step size of $2cm$ in the task space. Then we compute the joint configurations of the trajectory waypoints of these $3$ phases using IK. Finally, the interpolated excavation trajectory is sent to a position controller in both simulation and real world.

%\vspace{-5pt}
\section{Objects Excavation Learning and Planning}
\label{sec:excavation_learning}
In this section we present the design of our deep network models to predict the excavation success for rigid objects in clutter. We then propose an excavation planner leveraging the learned prediction excavation model.

% \liangjun{need to explain why classifier not the regression? how do you define a successful excavation, weight more than a threshold?}

\vspace{-5pt}
\subsection{Excavation Scene Representation}
\label{sec:exv_scene}
\begin{figure}[h]
    \centering
    \begin{subfigure}[b]{0.23\textwidth}
        \centering
        \includegraphics[width=\textwidth]{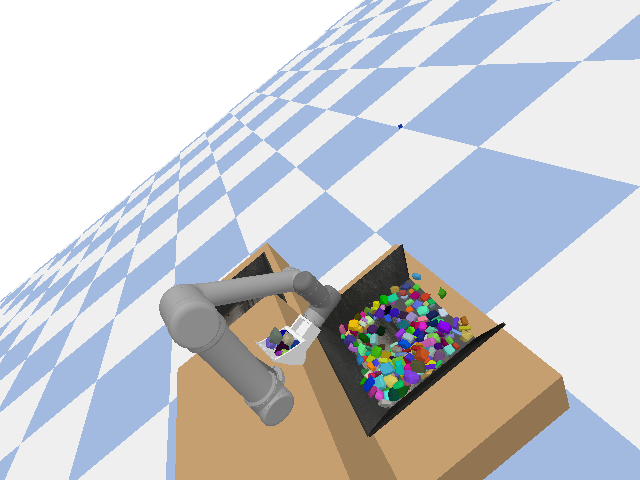}
        \caption{One RGB image of one excavation scene in simulation.}
        \label{fig:sim_rgb_example}
    \end{subfigure}
    \hfill
    \begin{subfigure}[b]{0.23\textwidth}   
        \centering 
        \includegraphics[width=\textwidth]{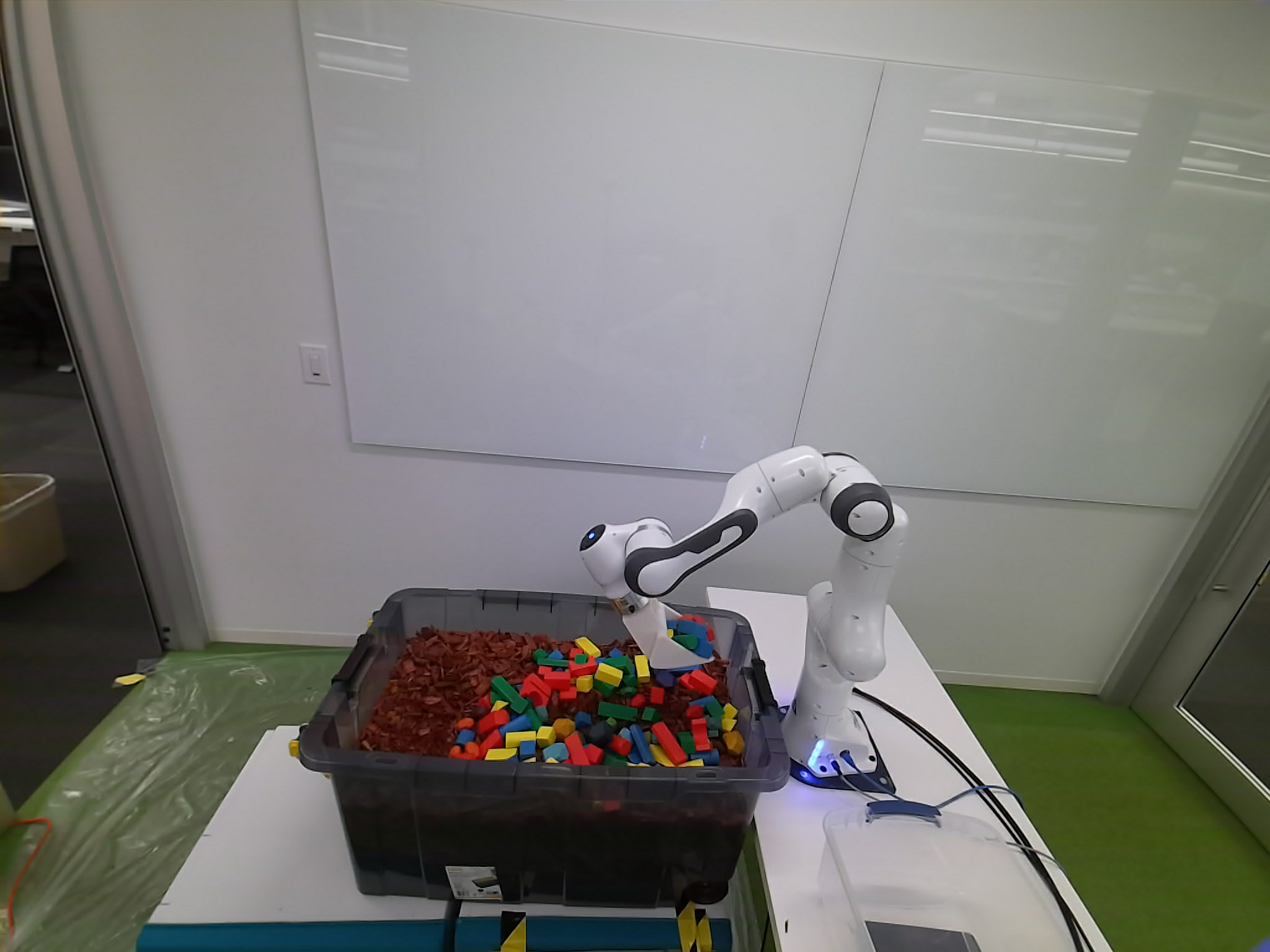}
        \caption{One Azure RGB image in real world.}  
        \label{fig:franka_exv_setup}
    \end{subfigure}
    \caption{The RGB image examples in simulation and real world.} 
    % \label{fig:rgb_example}
\end{figure}

We consider two visual representations for the excavation scene, RGBD image and voxel-grid. RGB and depth images are captured using the corresponding RGBD camera in simulation or real world. It turns out the RGBD image representation suffers from a large sim2real gap when transferring the learned excavation knowledge from simulation into real world, because (1) our simulated excavation environment (e.g., the geometry and color of the excavation tray and the color of the floor) differs from the real-world excavation environment; (2) the RGBD image depends on the camera intrinsics and extrinsics. Figure~\ref{fig:sim_rgb_example} and Figure~\ref{fig:franka_exv_setup} show the RGB image of the simulation and real world respectively. 

% To overcome the sim2real issue, we propose a 3D voxel-grid representation~\cite{lu2020multi, jetchev2010trajectory} of the excavation scene. 
% In~\cite{lu2020multi}, \hl{an voxel-based object representation and two 3D CNNs are presented for multi-fingered grasp learning and planning.}
% \hl{Jetchev and Toussaint model environments with voxel-grids and present a novel way for faster movement planning in such environments by predicting good path
% initializations}~\cite{jetchev2010trajectory}.

To overcome the sim2real issue, we propose a 3D voxel-grid representation of the excavation scene.
The pointcloud (i.e., depth) of the excavation scene obtained from the same RGBD camera is first transformed into the objects tray frame. Then we filter the transformed pointcloud according to the specific excavation cuboid space and voxelize the filtered pointcloud to generate its voxel-grid. More details of the excavation space specification for simulation and real world can be found in Section~\ref{sec:exv_scene_setup_sim} and~\ref{sec:exv_scene_setup_real} of the Appendix. The voxel-grid has a dimension of $64 \times 64 \times 32$ with a resolution of $0.01$ m. An example of the voxel-grid visualization and its source pointcloud are shown in Figure~\ref{fig:exv_voxel_net}.
The voxel-grid dimension and resolution are empirically designed to cover the excavation space and maintain a reasonable level of visual details, similar to the voxelization for grasping in~\cite{lu2020multi}.

Since our voxelization only focuses the specified excavation space, the voxel-grid representation is not affected by the environment surroundings. Moreover, our voxel-grid representation is agnostic to the camera intrinsics and extrinsics, because the voxelization is applied in the tray frame instead of the camera frame. Our experimental results in Section~\ref{sec:experiments} demonstrate the sim2real benefits of the voxel-grid representation over the RGBD one. 
%\liangjun{ move domain rand to future work} As another option, we would like to examine domain randomization~\cite{tobin2017domain} to shrink the sim2real gap of the RGBD representation in the future. 

%so that the representation becomes agnostic to the camera intrinsic, extrinsic and distortion parameters between simulation and real world. 
%In real world, the RGBD data can be captured using sensor, e.g. Azure Kinect camera. The main drawback of RGBD representation for excavation learning is the visual perception gap between the simulation and real world. Simulators are used for collecting a large set of training samples, while the model needs to be used for real world. To overcome this issue, we propose a voxel-grid representation of the scene so that the representation becomes agnostic to the camera intrinsic, extrinsic and distortion parameters between simulation and real world. 
% \liangjun{Describe how the voxel-grid representation is generated}

% domain randomization

\vspace{-5pt}
\subsection{Excavation Prediction Models}
% \liangjun{mention RGBD and Voxel. The text below is only for rgbd}. 
We model the excavation prediction as a binary classification problem. The excavation classifier predicts the probability of excavation success (i.e., bucket filling success), \(Y\), as a function of an excavation instance. 
% An excavation instance is defined to be the pair of the visual scene representation $Z$ and the excavation task trajectory $T$, as described in Section~\ref{sec:task_overview}. 
We propose two CNN models to predict the excavation success probability, namely ``excavation-RGBD-net" and ``excavation-voxel-net". Each model takes an excavation instance composed of a task trajectory and a RGBD image/voxel-grid as input and predicts the excavation success probability as output. 

ResNet~\cite{he2016deep} provides one of the state of the art CNN architectures for various computer vision tasks such as image classification and object detection. We choose ResNet-$18$ as the backbone architecture of excavation-RGBD-net and naturally extend ResNet-$18$ to 3D CNN as the backbone of excavation-voxel-net. The offline validation results in Section~\ref{sec:offline_eval} of the Appendix and the experiments in Section~\ref{sec:experiments} empirically show the effectiveness of both models, especially excavation-voxel-net. We believe other alternative network structures such as scoop \& dump-net in~\cite{schenck2017learning} for the 2D CNN RGBD model and voxel-config-net in~\cite{lu2020multi} or the shape completion CNN in~\cite{varley2017shape} for the 3D voxel model could also potentially work well.

%Our excavation-RGBD-net represents the given excavation scene using its RGBD image. Our excavation-voxel-net represents the given excavation scene as the voxel-grid of the excavation area.

\begin{figure}[h]
    \centering
    \includegraphics[width=0.48\textwidth]{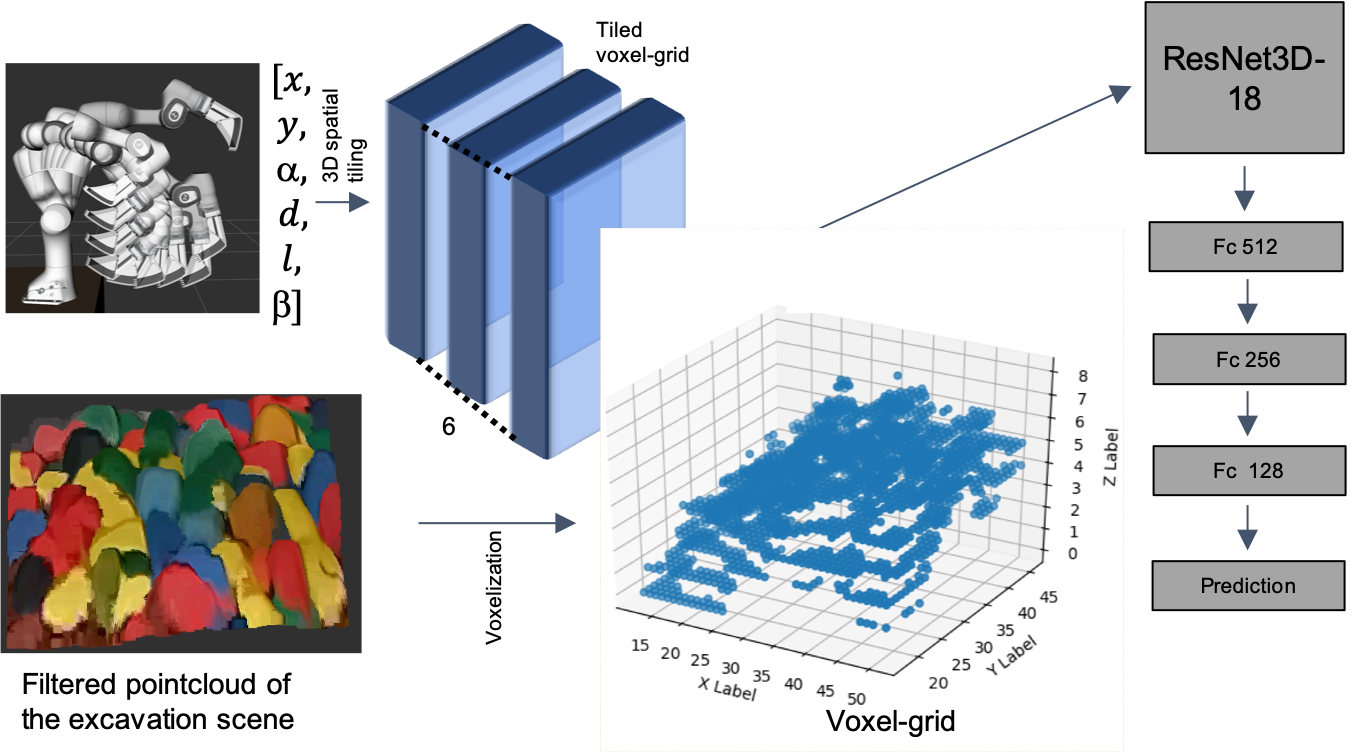}
    \caption{The excavation-voxel-net architecture.}
    \label{fig:exv_voxel_net}
\end{figure}

Figure~\ref{fig:exv_voxel_net} shows the architecture of our excavation-voxel-net using the ResNet-$18$ backbone.
We tile each trajectory parameter point-wise across the $64 \times 64 \times 32$ voxel-grid dimension (cf.~\cite{levine2018learning}). We then concatenate the tiled trajectory parameter voxel-grids with the scene voxel-grid to generate the final input voxel-grid of the given excavation instance. The input voxel-grid has $7$ channels (i.e., the dimension of each voxel) in total, including $1$ for the scene voxel-grid and $6$ for the tiled trajectory parameters.

%\liangjun{is this 3, since it is 3D. should $7$ be $10$} 
The 2D convolution filters of the raw ResNet-18 are replaced with 3D convolution filters to build the ResNet3D-18 backbone.
We feed the input voxel-grid into ResNet3D-18 to generate a $1000$-dimensional feature vector. The ResNet3D feature is then processed by $3$ fully-connected layers followed by a sigmoid output layer to generate the excavation success probability.
These $3$ fully-connected layers have $512$, $256$, and $128$ ReLu neurons, respectively. The design of the fully-connected layers is inspired from the voxel-config-net in~\cite{lu2020multi}, which is further tuned empirically during training.
We apply batch normalization for all fully-connected layers except the output layer. 
We train our the excavation classifier using the cross entropy loss. 

The excavation-RGBD-net shares a similar architecture with the excavation-voxel-net, except we use the raw ResNet-18 backbone with 2D convolution and tile the trajectory parameters in the image space instead of voxel-grid space.

% \textbf{TODO: describe the regressor.} We also tried to model the excavation prediction as a regression problem. However, we found the CNN regressor tends to overfit heavily on the offline testing set. 
% Since classification is about predicting a label and regression is about predicting a continuous quantity, we believe regression needs more training data to perform well. We plan to investigate the excavation regression prediction deeper in the future. 

To compare with classification, we also model the excavation prediction as a regression problem. Excavation-RGBD-net and excavation-voxel-net are adapted to ``excavation-RGBD-reg-net" and ``excavation-voxel-reg-net" respectively by replacing the sigmoid output layer with the fully-connected layer. The regression models are trained using the smooth L1 loss (i.e., Huber loss)\footnote{\url{https://pytorch.org/docs/stable/generated/torch.nn.SmoothL1Loss.html}}.

In order to show the importance of the scene dependency for excavation learning and provide a data-driven baseline for experiments, we also develop a fully-connected excavation classification network ``excavation-traj-net". The scene-independent excavation-traj-net only takes the task trajectory without the visual scene representation as input. It has $4$ fully-connected layers with $512$, $256$, and $128$ ReLu neurons respectively. Its final sigmoid layer outputs the excavation success probability.

In summary, five excavation prediction models are presented: excavation-RGBD-net, excavation-voxel-net, excavation-traj-net, excavation-RGBD-reg-net, and excavation-voxel-reg-net.

\vspace{-7pt}
\vspace{-5pt}

\subsection{Learning-based Excavation Planning}
\label{sec:exv_planning}
% \textbf{TODO: the optimization objective function}
Given the excavation scene visual representation $Z$, our goal is to plan an excavation trajectory $T$ that maximizes the the probability of excavation success, \(Y\). Similar with the grasp planner in~\cite{lu2017grasp}, we formulate the excavation planning as an optimization problem:
\begin{equation}
\begin{aligned}
& \underset{T}{\text{argmax}}
& & p(Y=1 | T, Z, W) = f(T, Z, W) \\
\end{aligned}
\label{eq:inf_obj}
\end{equation}
\vspace{-2pt}
In Eq.~\ref{eq:inf_obj} \(f(T, Z, W)\) defines a neural network classifier with logistic output trained to predict the excavation success probability as a Bernoulli distribution over \(Y\). The parameters $W$ define the neural network parameters.

% We propose planning grasps by finding the grasp configuration which maximizes the grasp success probability. We use a deep neural network to predict the probability of grasp success, \(Y\), as a function of the sensor readings, \(z\), and hand configuration, \(\theta\). We formalize probabilistic grasp inference as an optimization problem:
% \begin{equation}
% \begin{aligned}
% & \underset{\theta}{\text{argmax}}
% & & p(Y=1 | \theta, z, w) = f(\theta, z, w) \\
% & \text{subject to}
% & & {\theta}_{min} \leq \theta \leq {\theta}_{max}.
% \end{aligned}
% \label{eq:inf_obj}
% \end{equation}
% In Eq.~\ref{eq:inf_obj} \(f(\theta, z, w)\) defines a neural network classifier with logistic output trained to predict the grasp success probability as a Bernoulli distribution over \(Y\). The parameters $w$ define the neural network parameters. 

We use CEM~\cite{de2005tutorial} leveraging the learned excavation prediction model to solve the excavation optimization problem, similar with~\cite{schenck2017learning}. As a sampling-based optimization approach, CEM iteratively samples from the current distribution and selects the top $K$ samples using a scoring function to update the distribution.
% samples from the current distribution, sorts the samples using a scoring function, uses the top $K$ samples to update the  distribution and resample. 
We aim to optimize a Gaussian distribution of the $6$D task trajectory parameters, given the visual representation of the current excavation scene. $256$ heuristic excavation trajectories are generated using the random-heu excavation planner to initialize the Gaussian distribution. More details of the random-heu planner can be seen from Section~\ref{sec:heu_planner} of the Appendix. We have $5$ iterations for the CEM excavation planning. At each iteration, we first sample $256$ excavation trajectory samples from the current distribution. We predict the success probability of each sample using the learned excavation prediction network. Then we select the top $64$ samples with higher success probabilities to update the Gaussian distribution. To summarize, CEM uses the learned prediction model as a quality metric to iteratively improves the distribution of the task trajectory parameters through sampling and distribution updating. We sample $64$ excavation trajectories from the final CEM distribution, evaluate each one using the learned prediction model, and pick the one with the highest success probability, valid IK solution, and valid attacking point range as the planned task trajectory.

% Sampling-based planning leveraging a learned prediction model is a common approach for manipulation learning~\cite{lenz2015deep}. We also benchmark a sampling-based excavation planner in the experiment. We generate $100$ excavation trajectories using the heu-random planner for the current scene. Then we evaluate all these $100$ heu-random excavation trajectories using the trained excavation prediction network and select the trajectory with the highest success probability and valid IK as the final excavation plan. 

% \vspace{-5pt}
% \section{Data collection}
% \label{sec:data_training}
% \input{6_training_data_collection}

%\vspace{-5pt}
\section{Excavation Experiments}
\label{sec:experiments}
In this section, we first describe the excavation experiment setup and results in simulation. Then we introduce the experiment setup and results in real world. Our learning-based planners are compared with two heuristic planners and a data-driven baseline planner in simulation and real world. Our experimental results demonstrate the learning-based planners are able to plan high-quality excavations and significantly outperform the baseline methods. The data collection, model training, offline validation, more detailed results analysis, and ablation study are provided in the Appendix.

\vspace{-5pt}
\subsection{Experiment Setup in Simulation}
\label{sec:exp_setup_sim}

We collect the training data and perform simulated experiments in PyBullet\footnote{\url{https://pybullet.org/wordpress/}}. A UR5 robot arm is used for excavation data collection. 
The UR5 arm has 6 DoF in total. We control the shoulder panning, shoulder lifting, elbow, and the first wrist joints of the UR5 arm and disable the other two wrist joints by fixing their joint angles in simulation. 
% For the home pose of UR5 robot, the upper arm is roughly vertical, the forearm is roughly horizontal, and the excavation bucket is roughly vertical and points down towards the objects tray. 
A 3D designed bucket is used as the end-effector of UR5 in simulation. The full volume of the bucket is $450$ cm$^3$.

% We use an UR5 arm with a 3D designed bucket to perform excavation experiments in simulation. The full bucket volume of the bucket is $450$ cm$^3$. The experiment setup in simulation is the same with data collection in Section~\ref{sec:data_collection}.

\begin{figure}[h]
    \centering
    \begin{subfigure}[b]{0.23\textwidth}
        \centering
        \includegraphics[width=\textwidth]{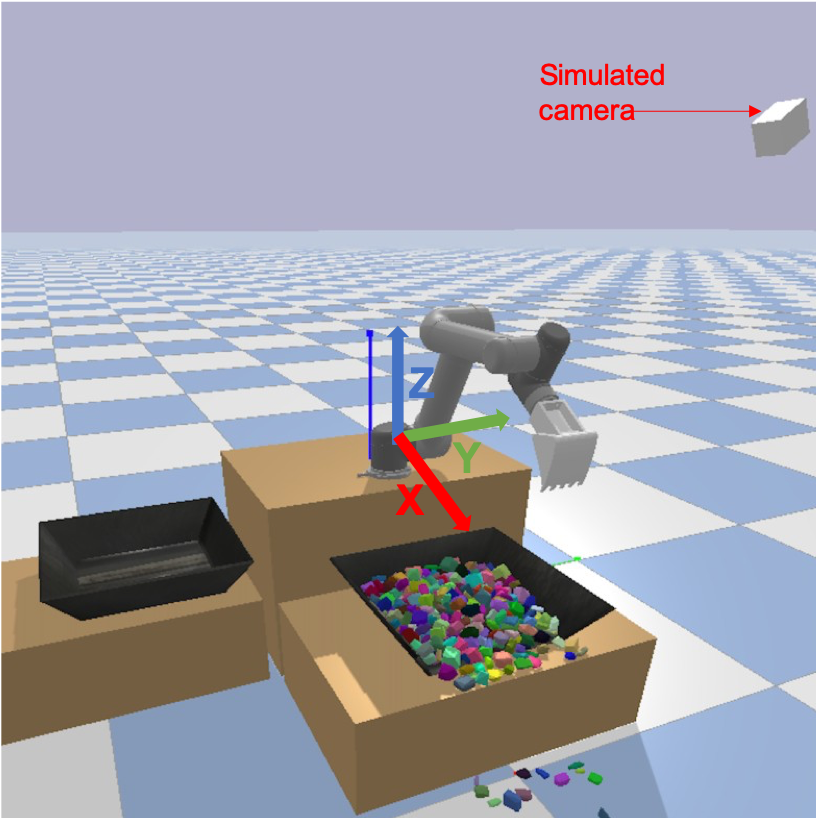}
        \caption{The RGBD camera setup in simulation.} 
        %The white box on the top right represents the camera.}
        \label{fig:sim_cam_setup}
    \end{subfigure}
    \hfill
    \begin{subfigure}[b]{0.23\textwidth}   
        \centering 
        \includegraphics[width=\textwidth]{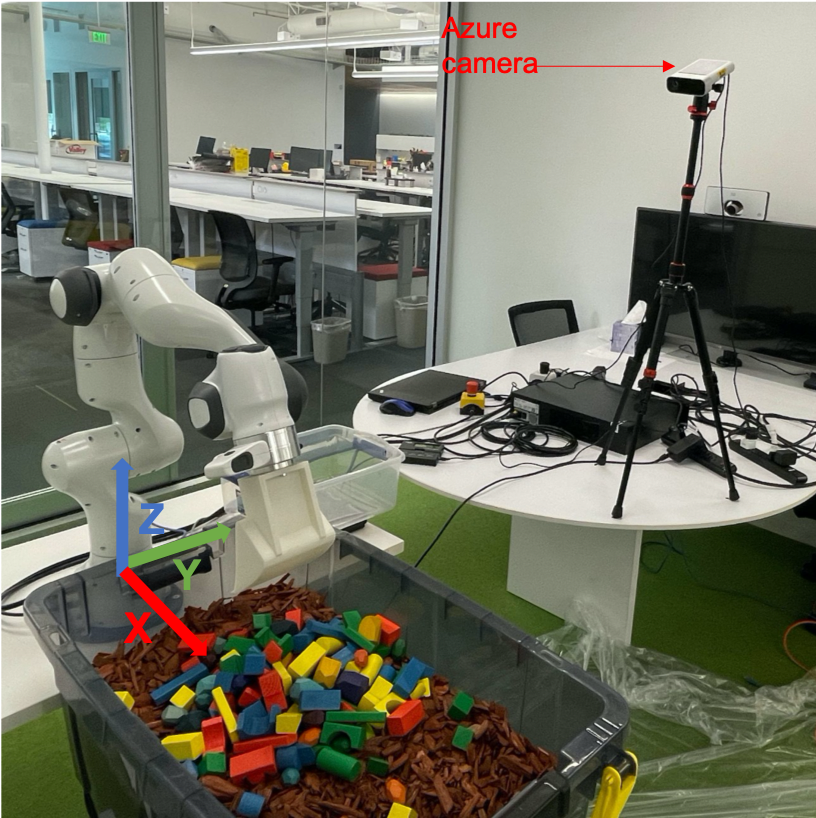}
        \caption{The Azure RGBD camera setup in real world.} % The white camera is mounted on the black tripod.}  
        \label{fig:real_cam_setup}
    \end{subfigure}
    \caption{The RGBD camera setup in simulation and real world.}
    % \label{fig:rgb_example}
\end{figure}

The RGB and depth image of each excavation trial are generated by the built-in simulated camera in PyBullet. Figure~\ref{fig:sim_cam_setup} shows the camera setup in simulation. One example of the RGB image generated by the simulated camera can be seen from Figure~\ref{fig:sim_rgb_example} in the Appendix. More details of the camera and excavation scene setup for real-robot experiments are discussed in Section~\ref{sec:exv_scene_setup_sim} of the Appendix.

% The camera locates at $(0.5$ m, $0.8$ m, $0.91$ m$)$ in the robot base frame. The $y$ axis of the robot frame points from the robot base to the tray center. Its $z$ axis is along the gravity direction.
% Both the transformation between the camera and the robot base frame and the transformation between the robot base frame and the tray frame are known. 
% With these two transformations, the pointcloud obtained from the PyBullet RGBD camera can be transformed into the tray frame for grid map and voxel-grid generation. 

%\vspace{-7pt}
\vspace{-5pt}

% \vspace{-7pt}
% \begin{figure}[h]
%     \centering
%     \begin{subfigure}[b]{0.23\textwidth}
%         \centering
%         \includegraphics[width=\textwidth]{figures/scene_0_sample_0_after_exv.png}
%         \caption{One RGB image of one excavation scene in simulation.}
%         \label{fig:sim_rgb_example}
%     \end{subfigure}
%     \hfill
%     \begin{subfigure}[b]{0.23\textwidth}   
%         \centering 
%         \includegraphics[width=\textwidth]{figures/franka_exv_setup.png}
%         \caption{One Azure RGB image in real world.}  
%         \label{fig:franka_exv_setup}
%     \end{subfigure}
%     \caption{The RGB image examples in simulation and real world.} 
%     % \label{fig:rgb_example}
% \end{figure}
% \vspace{-5pt}

% We sample the number of objects $n$ uniformly in the range of $[200, 400]$ for each excavation scene. We then spawn $n$ testing objects with random poses into the tray for the current excavation scene. The testing object meshes are unseen from training as described in Section~\ref{sec:data_collection}.
% The $0.38 \times 0.4 \times 0.3$ m$^3$ cuboid range is specified to filter the pointcloud of the excavation scene in the tray frame, which is then used for grid map and voxel-grid generation. This cuboid range covers the excavation space of rigid objects in the tray in simulation.

For each experiment trial of a certain excavation planner, the joint space trajectory is interpolated and computed from its planned task trajectory $T$ using IK, and sent the joint space waypoints to a joint position controller of the UR5 arm in simulation.

% \vspace{-5pt}
\subsection{Experiments in Simulation}
\label{sec:exp_sim}
% \textbf{Report the success probability mean and std of CEM-voxel and CEM-RGBD in simulation.}
% Explain the reason why less experiment excavations/more object number range for each scene than data collection. Explain the success threshold.\\

We perform simulated experiments to evaluate the learning-based planner of excavation-voxel-net, excavation-RGBD-net, excavation-voxel-reg-net, excavation-RGBD-reg-net, and excavation-traj-net. We name them ``CEM-voxel", ``CEM-RGBD", ``CEM-voxel-reg", ``CEM-RGBD-reg", and ``CEM-traj" respectively.
CEM-traj serves as a data-driven baseline planner without visual scene representation input.
In addition, these five learning-based planners are compared with two heuristic planners: random-heu and highest-heu. More details of these two heuristic planners can be found in Section~\ref{sec:heu_planner} of the Appendix. $100$ excavation episodes are experimented for each method. $10$ excavation trials are sequentially performed for each excavation episode. That gives us $1000$ excavation experimented trials in total for each method. 

The simulated results of all seven methods are presented in Table~\ref{table:exv_exp_sim_200_400}. We benchmark the excavations of each planner using three metrics: the volume of excavated objects (excavation volume), the excavated objects number, and the excavation success rate. Same as the model training in Section~\ref{sec:exv_train} of the Appendix, if the total volume of a sample's successfully excavated objects is above $134$ cm$^{3}$ (i.e., $30\%$ bucket filling rate), it is counted as a success, otherwise a failure. We also report the computation time of each planner. 

The mean with standard deviation in parentheses are listed for all metrics except the success rate. The mean and standard deviation for each method are computed across its $1000$ experimented excavation trials. 
% As we can see from Table~\ref{table:exv_exp_sim_200_400}, CEM-voxel is the best excavation planner in terms of the excavation volume, excavated object number, and success rate. CEM-voxel excavates objects of $159$ $cm^3$ per excavation in average, which is $35.3\%$ of the full bucket volume (i.e. bucket volume filling rate). 
As shown in Table~\ref{table:exv_exp_sim_200_400}, CEM-voxel achieves the best excavation performance in terms of the excavation volume, excavated object number, and success rate. CEM-voxel excavates objects of $136$ $cm^3$ per excavation in average, which is $30.2\%$ of the full bucket volume (i.e. bucket volume filling rate). 
% All five learning-based planners outperform these two heuristic planners in terms of these 3 excavation metrics. 
CEM-voxel, CEM-RGBD, and CEM-voxel-reg outperform the two heuristic planners and CEM-traj by relatively large margins in terms of these 3 excavation metrics, which shows the effectiveness of the scene-dependent excavation learning.

Classification-based CEM-voxel and CEM-RGBD perform better than regression-based CEM-voxel-reg and CEM-RGBD-reg respectively. Since classification is about predicting a label and regression is about predicting a continuous quantity, we believe excavation regression is more complex and needs a lot more training data to perform as well as or better than excavation classification.

The fact that scene-dependent planner CEM-voxel, CEM-RGBD, and CEM-voxel-reg significantly outperform the scene-independent CEM-traj planner demonstrates that it is important to learn to plan excavation trajectories based on the visual scene information.

The five learning-based planners all have higher standard deviations in terms of excavation volume and objects number than the two heuristic planners. CEM-voxel has the highest standard deviation. The experiment results of heuristic planners are dominated by failure excavations with low excavation volumes. Learning-based planners, especially CEM-voxel, generate excavations with relatively higher excavation volumes. This makes the excavation volume distribution of learning-based planners more uniform and have larger standard deviations, which is shown by the volume histogram of different planners in Figure~\ref{fig:hist_sim_200_400} of the Appendix. 
% \textbf{TODO: The CEM-voxel is also the least consistent (highest standard deviation). The authors should discuss the practical implications of this, as less consistent bucket filling could be the result of unpredictable digging behaviour. }

\begin{figure}[h]
    \centering
    \includegraphics[width=0.45\textwidth]{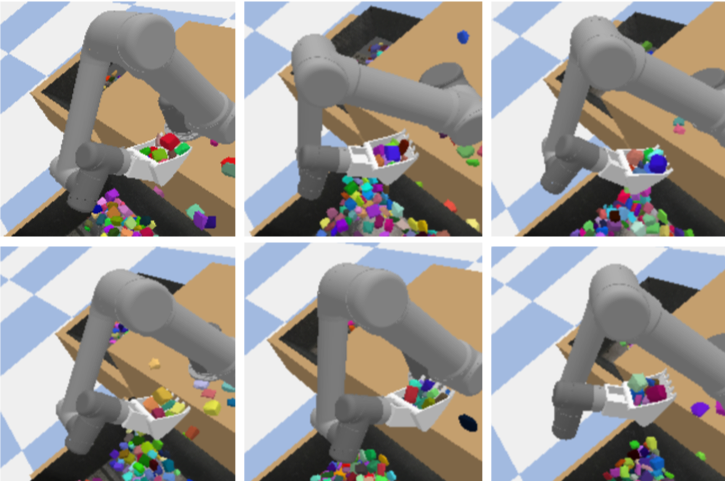}
    \caption{We show 6 high-quality simulated excavations generated by the CEM-voxel planner in this figure.}
    \label{fig:sim_exv_example}
\end{figure}

In terms of computation speed, heuristic planners spend $0.2$ second to plan one excavation trajectory. CEM-voxel, CEM-RGBD, CEM-voxel-reg, and CEM-RGBD-reg takes more than $10$ seconds to generate one excavation trajectory. It costs CEM-traj $3$ seconds to plan a trajectory. 
Finally, Figure~\ref{fig:sim_exv_example} visualizes $6$ high-quality excavation examples planned by the CEM-voxel planner in simulation. 

% Explain why heu-highest performs worse than heu-random?

% We plot of the histogram of the volume of excavated objects for the $1000$ experimented excavations of each method in Figure~\ref{fig:hist_sim_200_400}. The histogram also shows the learning-based planners can excavate objects with larger volumes than heuristic planners. Finally, Figure~\ref{fig:sim_exv_example} visualizes $6$ high-quality excavation examples planned by the CEM-voxel planner in simulation. 

% \vspace{-7pt}
% \begin{table}[h!]
% \begin{center}
% \resizebox{0.45\textwidth}{!}{
% \begin{tabular}{ |c|c|c|c|c| } 
%  \hline
% Method  & Volume ($cm^3$) & Number & Success Rate & Time (s)\\ 
%  \hline
%  CEM-voxel & $159.48$ $(120.33)$ & $8.78$ $(6.6)$ & $57.2\%$ & $10.47$ $(0.15)$\\
%  \hline
%  CEM-RGBD & $127.17$ $(107.65)$ & $7.24$ $(6.22)$ & $45.9\%$ & $17.85$ $(0.91)$\\
%  \hline
%  random-heu & $105.78$ $(102.02)$ & $5.82$ $(5.72)$ & $35.1\%$ & $0.2$ $(0.03)$ \\ 
%  \hline
%  highest-heu & $84.55$ $(87.97)$ & $4.29$ $(4.86)$ & $29\%$ & $0.2$ $(0.02)$ \\ 
%  \hline
% \end{tabular}
% }
% \caption{The experimental results of four excavation planners in simulation.}
% \label{table:exv_exp_sim_200_400}
% \end{center}
% \end{table}

% \vspace{-5pt}
\begin{table}[h!]
\begin{center}
\resizebox{0.45\textwidth}{!}{
\begin{tabular}{ |c|c|c|c|c| } 
 \hline
Method & Volume (cm$^3$) & Number & Success Rate & Time (s) \\ 
 \hline
 CEM-voxel & $136.23$ $(106.14)$ & $7.58$ $(6.05)$ & $51.9\%$ & $10.5$ $(0.77)$ \\
 \hline
 CEM-RGBD & $129.78$ $(101.50)$ & $7.51$ $(6.03)$ & $48.1\%$ & $17.3$ $(0.28)$ \\
 \hline
 CEM-voxel-reg & $127.89$ $(105.22)$ & $7.15$ $(6.09)$ & $47\%$ & $10.44$ $(0.29)$ \\
 \hline
 CEM-RGBD-reg & $107.93$ $(98.18)$ & $6.29$ $(5.73)$ & $35.3\%$ & $17.4$ $(0.27)$ \\
 \hline
 CEM-traj & $97.27$ $(100.73)$ & $5.54$ $(5.87)$ & $32.4\%$ & $3.17$ $(0.28)$ \\
 \hline
 random-heu & $85.81$ $(87.65)$ & $4.73$ $(4.99)$ & $28.4\%$ & $0.2$ $(0.03)$ \\ 
 \hline
 highest-heu & $67.24$ $(76.43)$ & $3.36$ $(4.19)$ & $19.3\%$ & $0.2$ $(0.02)$ \\ 
 \hline
\end{tabular}
}
\caption{The experimental results of seven excavation planners in simulation.}
\label{table:exv_exp_sim_200_400}
\end{center}
\end{table}
%\vspace{-5pt}

\vspace{-5pt}
\vspace{-7pt}

%#################################################################################################
%\vspace{-10pt}
\subsection{Experiment Setup in Real World}
%\vspace{-5pt}
Real-robot excavation experiments are performed using a Franka Panda robotic arm. The Franka Panda arm has 7 DOF in total. We control the shoulder panning, shoulder lifting, elbow lifting, and the wrist lifting joint of the Franka arm as the excavation joints and disable the other three joints (i.e. the elbow panning and the last two wrist joints) by fixing their joint angles. 
The same bucket model used in simulation is 3D printed as the Franka arm end-effector. 
The Azure Kinect camera generates the RGBD image and pointcloud of the excavation scene. Figure~\ref{fig:real_cam_setup} shows the camera setup in real world. 
An example of the Azure RGB image showing the excavation setup can be seen from Figure~\ref{fig:franka_exv_setup} in the Appendix. 
More details of the camera, robot, and excavation scene setup for real-robot experiments are introduced in Section~\ref{sec:exv_scene_setup_real} of the Appendix.

% For each experiment trial of a certain planner, we compute the joint space trajectory from the planned task trajectory $T$ using IK and send the joint space trajectory to the built-in joint position controller of the Franka arm.
% During penetration, the robot is automatically commanded to alternatively shift the bucket back or forth by $2$ cm horizontally per waypoint, which helps to prevent the robot getting stuck. We discuss more about the robot wiggling in Section~\ref{sec:exv_scene_setup_real} of the Appendix. 

For each experiment trial of a certain planner, we compute the joint space trajectory from the planned task trajectory $T$ using IK and send the joint space trajectory to the built-in joint position controller of the Franka arm.
Compared with hydraulic excavator arms, the Franka arm can only produce a limited amount of force and torque. For example, Franka's force and torque range along $z$ (i.e., gravity direction) are $[-50, 150]$ $N$ and $[-10, 10]$ $Nm$ respectively. Considering the large resistive force of rigid objects, this makes it hard for the bucket to penetrate into the rigid objects. 
During penetration, the robot is automatically commanded to alternatively shift the bucket back or forth by $2$ cm horizontally per waypoint, which helps to prevent the robot getting stuck.

% \begin{figure}[h]
%     \centering
%     \includegraphics[width=0.25\textwidth]{figures/franka_exv_setup.png}
%     \caption{The RGB image of one excavation example scene in real world generated by the Azure Kinect camera.}
%     %  We have the excavated objects in the white bucket. Objects are put into the excavation tray in front of the robot with random poses and colors. Excavated objects will be dumped into the dumping tray at the left of the robot after each excavation.
%     \label{fig:franka_exv_setup}
% \end{figure}

\vspace{-5pt}
\subsection{Real Robot Experiments}
The excavation model learned on UR5 in simulation is transferred to Franka in real world for rigid objects excavation experiments. The representation of the excavation trajectory in task space allows us to transfer the excavation prediction model from one hardware platform to another with similar kinematic reachability. The reachability of UR5 and Franka arm are $850$ mm and $800$ mm respectively. In addition to the task trajectory representation, we represent excavation poses in the tray frame to make excavation learning and planning agnostic to different tray poses across simulation and real world.

Excavation experiments are performed to evaluate our learning-based planners CEM-voxel and CEM-RGBD, which achieves the best performance in simulation experiments. We also compare our learning-based planners with these two heuristic planners random-heu and highest-heu. We experiment $5$ excavation episodes for each method in real world. We randomly reset the rigid objects for each excavation episode. We perform $5$ excavation trials for each excavation episode. That gives us $25$ excavation experimented trials in total for each method. 

Details of these two heuristic planners for simulation are described in Section~\ref{sec:heu_planner} of the Appendix.
Random parameter ranges of heuristic planners in real world are smaller than that in simulation. Because experiments with large heuristic ranges can be unsafe for human or robot. For example, relatively long dragging lengths cause collision with the tray. Moreover, the Franka arm can only produce a limited amount of force and torque, which makes it difficult to penetrate into the rigid objects with a depth larger than $5$ cm.
Specifically, we randomly generate the attacking excavation angle $\alpha$ and the closing angle $\beta$ in the range of $[-110, -70]$ and $[-110, -140]$ degree respectively in real world. We randomly generate the penetration depth $d$ and the dragging length $l$ in the range of $[0.02, 0.05]$ m and $[0.02, 0.06]$ m respectively. The trajectory parameter range of heuristic planners also affect our learning-based planners, since we generate heuristic excavation trajectories to initialize CEM, as described in Section~\ref{sec:exv_planning} of the Appendix.

The real-robot experiment results of all four methods are presented in Table~\ref{table:exv_exp_real}.
We evaluate the excavation performance in terms of the volume of excavated objects and the excavation success rate. The mean with standard deviation in parentheses are reported for the volume of excavated objects. The success threshold of the volume of excavated objects is $134$ cm$^3$, same as simulation. 

We also show the valid rate of each planner in the table. An excavation trial is treated as valid if the trajectory can be planned and executed successfully. Invalid excavation trials are mostly caused by limit exceeding of the robot force/torque. Large resistive force during excavation, especially penetration, and collision with the tray can both lead to the force/torque limit exceeding.
Examples of the Franka arm getting stuck due to force/torque limit exceeding are shown in Figure~\ref{fig:franka_stop_example} of the Appendix. 
In the future, we are interested to examine force control for excavation trajectory execution in order to mitigate force/torque limit exceeding.
We also count the excavation trials without valid trajectory IK as invalid.

As shown in Table~\ref{table:exv_exp_real}, the CEM-voxel planner significantly outperforms these other 3 planners in terms of the volume of excavated objects and success rate in real world. CEM-voxel excavates objects of $110$ cm$^3$ per excavation in average, which is $24.4\%$ of the full bucket volume. 
CEM-voxel significantly outperforms these two heuristic planners, which demonstrates the effectiveness of excavation learning in real world.
The fact that CEM-voxel outperforms CEM-RGBD shows that the voxel-based visual representation handles the sim2real gap better than the RGBD representation.
The computation time of each planner in the real world is similar with simulation.

The CEM-RGBD planner performs poorly in the real world, worse than random-heu and roughly on par with highest-heu. The attacking poses of the trajectories planned by CEM-RGBD are mostly close to the edge of the tray, which leads to invalid excavation trials with collision.
This is because the RGBD image representation suffers from a large sim2real gap when transferring the excavation knowledge gained in simulation into real world. In addition to the poor excavation performance, another evidence of the RGBD sim2real gap is the predicted success probabilities of the CEM-RGBD trajectories are close to zero in real world.
More details of the excavation scene visual representation are discussed in Section~\ref{sec:exv_scene}. 

\begin{figure}[h]
    \centering
    \includegraphics[width=0.45\textwidth]{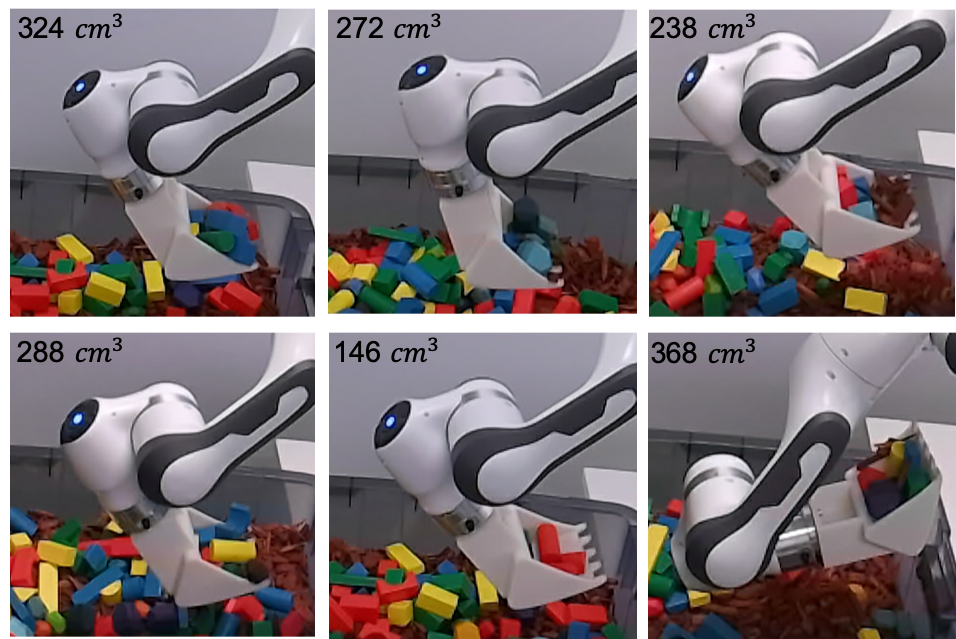}
    \caption{We show 6 successful real robot excavations generated by the CEM-voxel planner together with their bucket filling volumes in this figure.}
    \label{fig:franka_exv_example}
\end{figure}

In Figure~\ref{fig:franka_exv_example}, we show $6$ high-quality excavation examples planned by the CEM-voxel planner on the real robot. We also annotate the volume of excavated objects of each example.

%\vspace{-3pt}
\begin{table}[h!]
\begin{center}
\resizebox{0.45\textwidth}{!}{
\resizebox{0.45\textwidth}{!}{
\begin{tabular}{ |c|c|c|c|c| } 
 \hline
Method & Volume (cm$^3$) & Success Rate & Valid Rate \\ 
 \hline
 CEM-voxel & $110.32$ $(120.42)$ & $11/25$ & $17/25$ \\
 \hline
 CEM-RGBD & $13.84$ $(67.8)$ & $1/25$ & $2/25$ \\
 \hline
 random-heu & $50.24$ $(79.95)$ & $5/25$ & $12/25$ \\
 \hline
 highest-heu & $13.36$ $(26.51)$ & $0/25$ & $19/25$ \\
 \hline
\end{tabular}
}
}
\caption{The real-robot experimental results of four excavation planning methods.}
\label{table:exv_exp_real}
\end{center}
\end{table}
\vspace{-10pt}

\vspace{-5pt}
\vspace{-7pt}

% \begin{figure}[h]
%     \centering
%     \includegraphics[width=0.48\textwidth]{figures/exv_stop_example.png}
%     \caption{Three examples of the robot getting stuck due to force/torque limit exceeding during excavation.}
%     \label{fig:franka_stop_example}
% \end{figure}
% \vspace{-20pt}

%\vspace{-5pt}
\section{Conclusion}
\label{sec:discussion}
In conclusion, we propose multiple deep networks for success prediction of a new task, rigid objects excavation in clutter. We solve excavation planning as an optimization problem leveraging the learned prediction models. Our excavation experiments in simulation and real world show that our learning-based planner is able to generate high-quality excavations. The experimental results also demonstrate the advantage of the learning-based excavation planner over two heuristic planners and one data-driven scene-independent planner. 

We plan an excavation trajectory greedily by maximizing the excavation volumes of the current excavation. In the future, we would like to consider the long-term expected excavation reward of sequential excavations and investigate deep reinforcement learning for rigid objects excavation. We also plan to use force control instead of position control to make the excavation trajectory execution smoother and more robust to large resistive forces. Finally, we want to transfer the learning-based planner from robotic arms to real excavators.

%\vspace{-5pt}
%%%%%%%%%%%%%%%%%%%%%%%%%%%%%%%%%%%%%%%%%%%%%%%%%%%%%%%%%%%%%%%%%%%%%%%%%%%%%%%%
%\section*{APPENDIX}
%
%Appendixes should appear before the acknowledgment.

%\section*{ACKNOWLEDGMENT}

%\begin{acknowledgement}
%\end{acknowledgement}

\bibliographystyle{IEEEtran}
\bibliography{references}  % .bib

\clearpage
% \section*{APPENDIX}
\section{APPENDIX}
In this Appendix section, we first introduce the excavation scene setup in simulation and real world. Then we describe the data collection and training of the excavation prediction model, before presenting the offline evaluation of the learned model results. We also show the excavation volume histograms, trajectory analysis, and ablation experiments. 

As defined in Section~\ref{sec:task_traj}, a task space excavation trajectory $T$ contains $6$ parameters $T=(x, y, \alpha, d, l, \beta)$. We name $(x, y)$ and $(\alpha, d, l, \beta)$ of a task trajectory $T$ as ``Point of Attack" (PoA) and ``Geometric Trajectory Parameters" (GTP) respectively for the ablation study.

\vspace{-5pt}
%\subsection{Data collection}
\label{sec:data_training}
% In this section, we first introduce our data collection pipeline. Then we describe the training of the excavation prediction model, before showing the offline evaluation of the learned model results.

\subsection{Camera and Excavation Scene Setup in Simulation}
\label{sec:exv_scene_setup_sim}

The camera is located at $(0.5$ m, $0.8$ m, $0.91$ m$)$ in the robot base frame in simulation.
The robot frame and the camera location for simulation are shown in Figure~\ref{fig:sim_cam_setup}. 
% The $x$ axis of the robot frame points from the robot base to the tray center. The $z$ axis of the robot frame is along the gravity direction.
Both the transformation between the camera and the robot base frame and the transformation between the robot base frame and the tray frame are known. 
With these two transformations, the pointcloud obtained from the PyBullet RGBD camera can be transformed into the tray frame for grid map and voxel-grid generation.

We sample the number of objects $n$ uniformly in the range of $[200, 400]$ for each excavation scene. We then spawn $n$ testing objects with random poses into the tray for the current excavation scene. The testing object meshes are unseen from training as described in Section~\ref{sec:data_collection} of the Appendix.
The $0.38 \times 0.4 \times 0.3$ m$^3$ cuboid range is specified to filter the pointcloud of the excavation scene in the tray frame, which is then used for grid map and voxel-grid generation. This cuboid range covers the excavation space of rigid objects in the tray in simulation.

%\vspace{-7pt}
%\vspace{-5pt}

\subsection{Camera and Excavation Scene Setup in Real World}
\label{sec:exv_scene_setup_real}
The Azure camera is located at $(0.57$ m, $1$ m, $1.14$ m$)$ in the robot frame in real world. The robot frame and camera location in real world can be seen in Figure~\ref{fig:real_cam_setup}.
% The $y$ axis of the robot frame points from the robot base to the tray center. Its $z$ axis is along the gravity direction.
% An example of the Azure RGB image showing the excavation setup can be seen from Figure~\ref{fig:franka_exv_setup}. 
The transformation between the camera and the robot base frame is manually calibrated using an ArUco marker~\footnote{\url{https://docs.opencv.org/master/d5/dae/tutorial_aruco_detection.html}}. 
We manually define the tray frame with respect to the robot base frame according to the excavation range, which gives the transformation between the robot and the tray frame. The tray frame has the same orientation with the robot base frame. Its origin is defined to be the center of the excavation cuboid range.
Knowing these two transformations, the pointcloud obtained from the Azure camera can be transformed into the tray frame for grid map and voxel-grid generation.

\begin{figure}[h]
    \centering
    \includegraphics[width=0.48\textwidth]{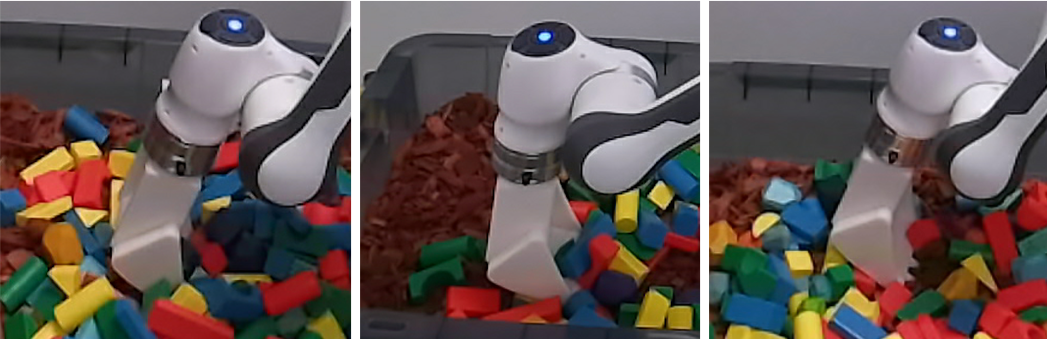}
    \caption{Three examples of the robot getting stuck due to force/torque limit exceeding during excavation.}
    \label{fig:franka_stop_example}
\end{figure}

$132$ rigid wooden objects with various geometrical shapes and colors are used for real robot experiments, including $100$ ``Melissa \& Doug wood blocks" and $32$ ``Biubee wooden stone balancing blocks". For example, there are objects with both convex and concave shapes. The density of the wooden rigid objects is estimated to be $0.5$ kg$/$cm$^3$. All of these rigid objects are unseen from the training. 

A layer of rocks with heavy mass is first put into the excavation tray, which stabilizes the tray during excavation. Then we lay a layer of red mulch on top of rocks. Finally rigid wooden objects are put on top of the red mulch in front of the Franka arm for excavation. We use the relatively deformable red mulch as the excavation surface for safety reasons. The $0.4 \times 0.3 \times 0.3$ m$^3$ cuboid range is specified to filter the pointcloud of the excavation scene in the tray frame, which is used for grid map and voxel-grid generation. This cuboid space covers the excavation space of the real-robot experiments. Roughly only the half of the tray space that is closer to the robot base is used for excavation experiments. 

An excavation episode is created by shaking these $132$ rigid objects in a box, and then pouring them into the excavation area of the tray. The robot dumps the excavated objects into a dumping tray after excavation for each trial. 
A certain amount of red mulch under the rigid objects can be excavated and dumped sometimes. On average the amount of excavated red mulch is relatively small across all experimented trials.

The desired dumping pose of the bucket end-effector is specified to be the center of the dumping tray. The robot first moves to the desired dumping pose, then pour the objects into the dumping tray by controlling the bucket to point down vertically. We use a kitchen scale under the dumping tray to weigh the objects dumped into the tray for each trial. Having the mass of the excavated objects and the objects density, we can compute the volume of excavated objects. 

% The Franka arm can only produce a limited amount of force and torque. For example, Franka's force and torque range along $z$ (i.e., gravity direction) are $[-50, 150]$ $N$ and $[-10, 10]$ $Nm$ respectively. Considering the large resistive force of rigid objects, this makes it hard for the bucket to penetrate into the rigid objects. 
% During penetration, the robot is automatically commanded to alternatively shift the bucket back or forth by $2$ cm horizontally per waypoint, which helps to prevent the robot getting stuck.

% \vspace{-5pt}
\subsection{Data Collection in Simulation}
\label{sec:data_collection}

% We collect the training data in PyBullet\footnote{\url{https://pybullet.org/wordpress/}} simulation environment. We use a UR5 robot arm for excavation data collection. 
% The UR5 arm has 6 DoF in total. We control the shoulder panning, shoulder lifting, elbow, and the first wrist joints of the UR5 arm and disable the other two wrist joints by fixing their joint angels for our excavation data collection and experiments in simulation. 
% % For the home pose of UR5 robot, the upper arm is roughly vertical, the forearm is roughly horizontal, and the excavation bucket is roughly vertical and points down towards the objects tray. 
% A 3D designed bucket is used as the end-effector of UR5 in simulation. The full bucket volume of the bucket is $450$ cm$^3$.

We use an UR5 arm with a 3D designed bucket to perform excavation experiments in simulation. The full bucket volume of the bucket is $450$ cm$^3$. The data collection setup is the same with the simulated experiment setup in Section~\ref{sec:exp_setup_sim}. The camera and excavation scene setup for data collection is described in Section~\ref{sec:exv_scene_setup_sim} of the Appendix. 
Rigid object meshes with random geometry for simulated excavation are generated using trimesh\footnote{\url{https://github.com/mikedh/trimesh}}. 
The number of vertices for each object mesh are randomly selected in the range of $10$ to $50$. The maximum value of each coordinate is uniformly sampled from $1$ cm to $5$ cm for the object mesh. The 3D coordinates of all vertices of the object mesh are randomly generated from the range of $0$ to its maximum coordinate values. We then compute the convex hull of the original mesh and use that as the final object mesh. We assume the object density to be $6$ g$/$cm$^3$ in simulation. We separately generate $100$k training and testing candidate object meshes. The training object mesh dataset is used for training data collection in simulation. The testing object mesh dataset are used for excavation prediction model offline evaluation and experiments in simulation.

A certain number of object meshes are randomly selected from the training objects set for each excavation episode of the data collection. We then spawn each selected object into the excavation tray with a random generated pose. The objects number of each scene is randomly and uniformly generated in the range of $50$ to $400$. $20$ excavation trials (i.e., samples) are sequentially executed for each excavation episode. We randomly select one of our two heuristic planners to use and compute the total volume of the objects excavated successfully at each trial.
% The RGB and depth image of each excavation trial are generated by the built-in simulated camera in PyBullet. One example of the simulated RGB image can be seen from Figure~\ref{fig:sim_rgb_example}. 
The excavated objects are dumped into the dumping tray for each excavation trial. We collected $50,000$ training excavation samples and $10,000$ testing samples for offline validation of the excavation prediction network.

\subsubsection{Heuristic Excavation Planners}
\label{sec:heu_planner}
We design two heuristic excavation planners for data collection, namely ``heu-random'' and ``heu-highest''. For the heu-random planner, we randomly select a grid map cell of the excavation scene and use its center as the 2D $(x, y)$ coordinate of the attacking excavation pose. For the heu-highest planner, we generate the 2D $(x, y)$ coordinate of the attacking excavation pose as the center of the grid map cell with the maximum height. We assume the attacking point is on the object clutter surface. Under this assumption, the $z$ coordinate value of the attacking excavation pose is computed as the height of the corresponding grid map cell.

The excavation grid map is generated from the point cloud of the excavation space using the grid map library in~\cite{Fankhauser2016GridMapLibrary} in both simulation and real world. 
% Figure~\ref{fig:grid_map_azure} and~\ref{fig:grid_map_sim} respectively visualizes one $(x, y, z)$ attacking point example of the heu-highest and heu-random planner.
We randomly generate the attacking excavation angle $\alpha$ and the closing angle $\beta$ in the range of $[-120, -60]$ and $[-180, -120]$ degree respectively. We randomly generate the penetration depth $d$ and the dragging length $l$ in the range of $[0.05, 0.2]$ m and $[0.05, 0.4]$ m respectively. The same random parameter ranges are used for data collection and experiments in simulation.

\vspace{-5pt}
\subsection{Excavation Prediction Model Training}
\label{sec:exv_train}
We collected $50,000$ training excavation samples in simulation. $45,000$ training samples are used for training and these other $5000$ training samples are used as the validation set. 
%The excavation prediction is learned as a binary classification problem in our work. 
For the excavation binary classification, if the total volume of a sample's successfully excavated objects is above $134$ cm$^3$ (i.e., $30\%$ bucket filling rate), the excavation sample is treated as a success, otherwise a failure. Excavation samples without valid task trajectory IK are labeled as failure excavations, using which we aim to learn to plan excavation trajectories with valid IK. $4768$ out of these $50,000$ ($10\%$) training samples are successful excavations.

% (i.e. $4$ spheres with $2$ cm radius)

We train all five excavation prediction models, including excavation-RGBD-net, excavation-voxel-net, excavation-RGBD-reg-net, excavation-voxel-reg-net, and excavation-traj-net, using the same specifications. 
In order to overcome the class imbalance (i.e., i.e. low percentage of successful excavation samples), the successful samples are oversampled to make the number of positive and negative samples roughly the same in each training epoch for all five models.
We compare training the excavation-RGBD-net from scratch and fine-tuning ResNet-$18$. We find training from scratch has significantly better performance, which we believe is because our excavation task is significantly different from the ResNet ImageNet classification. In addition to excavation-RGBD-net, we also train all other four models from scratch.

All networks are trained using the Adam optimizer with mini-batches of size $64$ for $50$ epochs. The learning rate starts at $0.1$ and decreases by $10\times$ every $10$ epochs. The training of excavation-RGBD-net and excavation-RGBD-reg-net take around $810$ minutes on an Alienware desktop computer with an Intel i7-6800K processors, 32GB RAM, and a Nvidia GeForce GTX TITAN Z graphics card. Excavation-voxel-net and excavation-voxel-reg-net take around $500$ minutes to train on the same machine. It takes excavation-traj-net around $100$ minutes to train on the same machine. We implement all excavation prediction networks in PyTorch.
We have released the data and the trained models in this link: \url{https://drive.google.com/drive/folders/1X54doBlf3QBjjNFTAZA48igO9VInY0k2?usp=sharing}

% models~\footnote{\url{https://drive.google.com/drive/folders/1X54doBlf3QBjjNFTAZA48igO9VInY0k2?usp=sharing}}.

\subsection{Excavation Prediction Model Offline Evaluation}
\label{sec:offline_eval}
We collected $10,000$ testing samples using the testing objects dataset in simulation for offline validation of the excavation prediction models. Among these $10,000$ testing samples, $967$ samples are successful excavations.

Table~\ref{table:offline_eval} shows the accuracy, precision, recall, and F$1$ score of three methods. The second, third, and forth row show the offline testing result of the excavation-voxel-net, excavation-RGBD-net, and excavation-traj-net respectively. The ``random-$0.5$" method in the fifth row refers to random guessing with a probability of $0.5$ for positive prediction. The ``random-$0.1$" method in the last row means random guessing with a probability of $0.1$ for positive prediction. The prediction metrics of random guessing show the classification challenges due to the low percentage of successful excavation samples. Excavation-voxel-net and excavation-RGBD-net perform reasonably well and significantly outperform random guessing in terms of these offline evaluation metrics. Excavation-voxel-net achieves the best offline evaluation performance. Excavation-traj-net performs worse than excavation-voxel-net and excavation-RGBD-net for the offline evaluation, but significantly better than random guessing. 

%\vspace{-7pt}
\begin{table}[h!]
\begin{center}
\resizebox{0.4\textwidth}{!}{
\begin{tabular}{ |c|c|c|c|c| } 
 \hline
Method & Accuracy & Precision & Recall & F$1$ \\ 
 \hline
 excavation-voxel-net & $0.904$ & $0.502$ & $0.637$ & $0.562$ \\
 \hline
 excavation-RGBD-net & $0.877$ & $0.399$ & $0.542$ & $0.459$ \\ 
 \hline
 excavation-traj-net & $0.731$ & $0.241$ & $0.827$ & $0.373$ \\ 
 \hline
 random-$0.5$ & $0.5$ & $0.1$ & $0.5$  & $0.17$ \\ 
 \hline
 random-$0.1$ & $0.82$ & $0.1$ & $0.1$ & $0.1$ \\ 
 \hline
\end{tabular}
}
\caption{The offline evaluation results of the excavation-voxel-net, excavation-RGBD-net, excavation-traj-net, and random guessing for classifying the excavation success on the testing set.}
\label{table:offline_eval}
\end{center}
\end{table}

% voxel regression
% volume_err 31.94, zero_volume_err 28.21, zero_volume_err 34.64
% volume_err std 37.66, zero_volume_err std 19.81, zero_volume_err std 46.31

% RGBD regression
% volume_err 35.22, zero_volume_err 27.05, zero_volume_err 41.14
% volume_err std 43.71, zero_volume_err std 30.73, zero_volume_err std 50.28

The excavation regression model excavation-voxel-reg-net and excavation-RGBD-net are also offline evaluated on the testing set using the L1-norm error. The mean and standard deviation of the testing L1-norm error of excavation-voxel-reg-net are $31.94$ cm$^3$ and $37.66$ cm$^3$ respectively. The testing L1-norm error of excavation-RGBD-reg-net has a mean of $35.22$ cm$^3$ and a standard deviation of $43.71$ cm$^3$. Both regression models achieve reasonably good testing performance.

\subsection{Excavation Volume Histograms of Experiments in Simulation}
%\textbf{TODO: explain why only sim histograms}

\begin{figure}[h]
    \centering
    \hfill
    \begin{subfigure}[b]{0.23\textwidth}   
        \centering 
        \includegraphics[width=\textwidth]{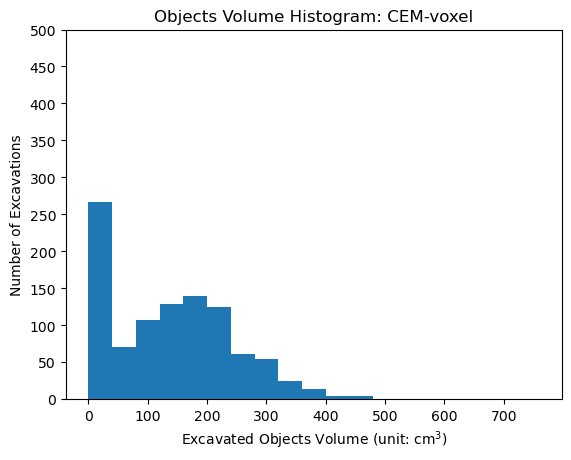}
        \caption{The excavation volume histogram of CEM-voxel.}  
        \label{fig:hist-cem-voxel}
    \end{subfigure}
    \begin{subfigure}[b]{0.23\textwidth}
        \centering
        \includegraphics[width=\textwidth]{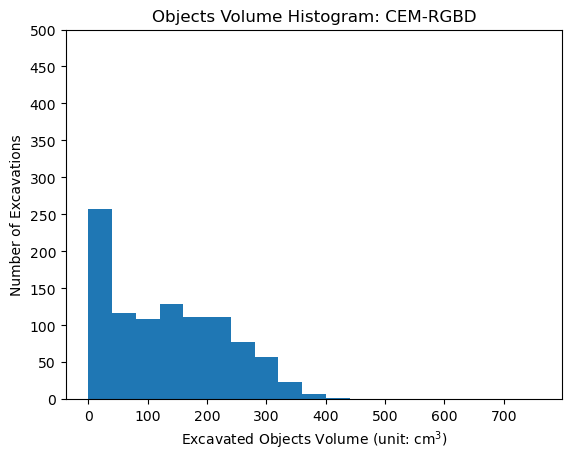}
        \caption{The excavation volume histogram of CEM-RGBD.}
        %\label{fig:}
    \end{subfigure}
    \vskip\baselineskip
    \hfill
    \begin{subfigure}[b]{0.23\textwidth}   
        \centering 
        \includegraphics[width=\textwidth]{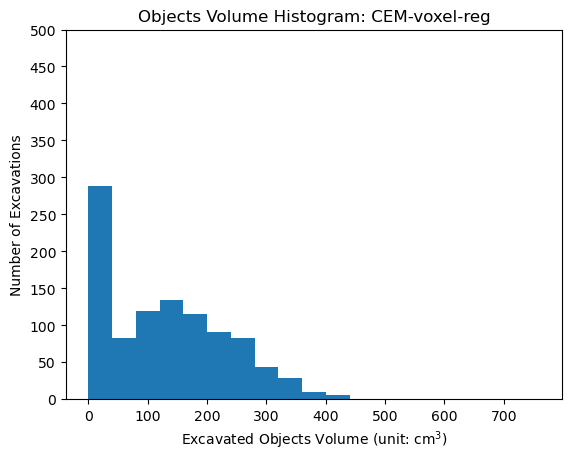}
        \caption{The excavation volume histogram of CEM-voxel-reg.}  
        %\label{fig:x}
    \end{subfigure}
    \begin{subfigure}[b]{0.23\textwidth}  
        \centering 
        \includegraphics[width=\textwidth]{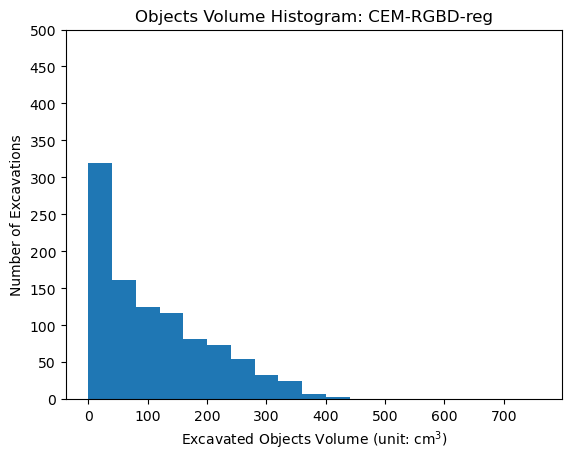}
        \caption{The excavation volume histogram of CEM-RGBD-reg.} 
        %\label{fig:x}
    \end{subfigure}
    \vskip\baselineskip
    \hfill
    \begin{subfigure}[b]{0.23\textwidth}   
        \centering 
        \includegraphics[width=\textwidth]{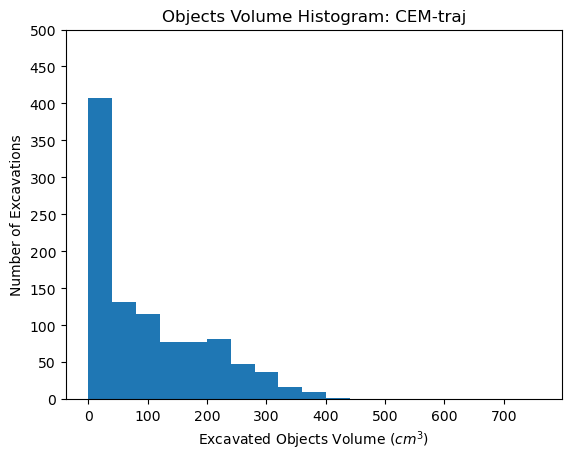}
        \caption{The excavation volume histogram of CEM-traj.}  
        %\label{fig:x}
    \end{subfigure}
    \begin{subfigure}[b]{0.23\textwidth}   
        \centering 
        \includegraphics[width=\textwidth]{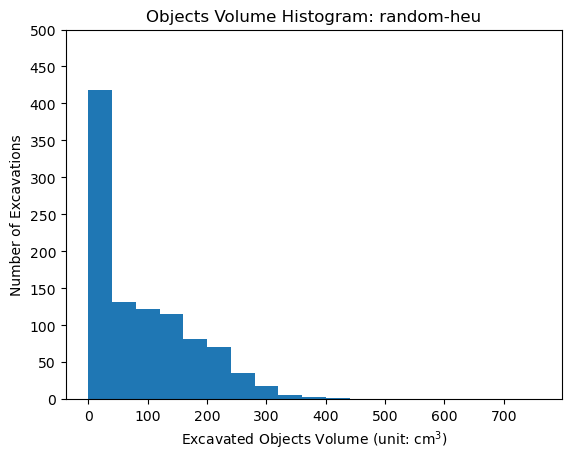}
        \caption{The excavation volume histogram of random-heu.}  
        \label{fig:hist-random-heu}
    \end{subfigure}
    \vskip\baselineskip
    \hfill
    \begin{subfigure}[b]{0.23\textwidth}   
        \centering 
        \includegraphics[width=\textwidth]{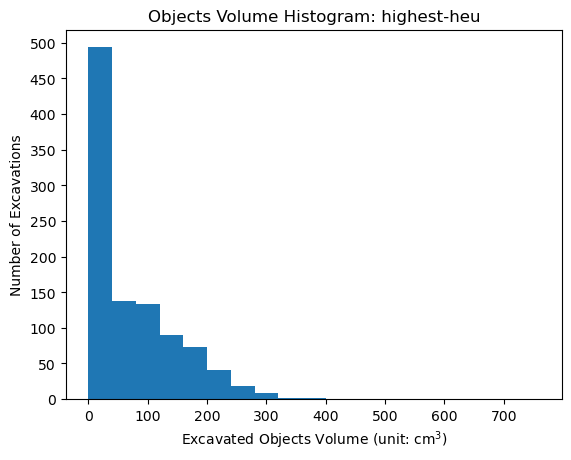}
        \caption{The excavation volume histogram of highest-heu.}  
        \label{fig:hist-highest-heu}
    \end{subfigure}
    \begin{subfigure}[b]{0.23\textwidth}  
        \centering 
        \includegraphics[width=\textwidth]{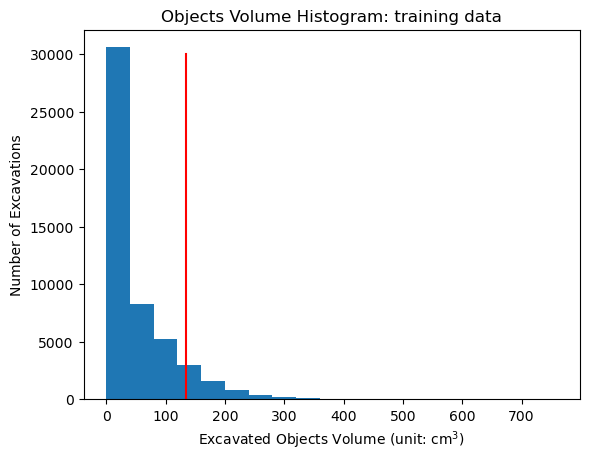}
        \caption{The excavation volume histogram of the training data.} 
        \label{fig:hist-training}
    \end{subfigure}
    % \begin{subfigure}[b]{0.24\textwidth}  
    %     \centering 
    %     \includegraphics[width=\textwidth]{figures/obj_volume_hist_training-data.png}
    %     \caption{The excavation volume histogram of the training data.} 
    %     \label{fig:hist-training}
    % \end{subfigure}
    % \caption{Figure~\ref{fig:hist-cem-voxel}-~\ref{fig:hist-highest-heu} show the histogram of the excavation volume of seven planners in simulated experiments. The $x$ axis represents the object volume. The $y$ axis represents the excavation volume.} 
    \caption{Figure \textbf{(a)}-\textbf{(g)} show the histogram of the excavation volume of seven planners in simulated experiments. The excavation volume histogram of the training data is shown in Figure \textbf{(h)}. The $x$ axis represents the excavation volume. The $y$ axis represents the number of excavations.} 
    \label{fig:hist_sim_200_400}
\end{figure}

The histograms of the excavation volume for the $1000$ simulated experimented excavations of seven planners are visualized in Figure~\ref{fig:hist-cem-voxel}-\ref{fig:hist-highest-heu}. 
In addition to the excavation volume means and the excavation rates in Table~\ref{table:exv_exp_sim_200_400}, the histograms further shows the learning-based planners excavate objects with larger volumes than heuristic planners. 
The histograms also show the distributions of learning-based planners are more uniform and have larger standard deviations than heuristic planners. 
Only the excavation histograms of simulation experiments are shown here, since the number of excavations in real world experiments is relatively small.

The excavation volume histogram of the training data is plotted in Figure~\ref{fig:hist-training}. We consider an excavation as a success if its excavation volume is above $134$ cm$^3$. The red vertical line in Figure~\ref{fig:hist-training} show where the excavation volume is $134$ cm$^3$.
With $134$ cm$^3$ as the excavation success threshold, $4768$ out of these $50,000$ ($10\%$) training samples are successful. 

Since there are a lot less successful training excavation samples than failure ones, we oversample the successful samples overcome this class imbalance issue in the excavation training, as described in Section~\ref{sec:exv_train} of the Appendix. Increasing the threshold to be larger than $134$ cm$^3$ will lead to even less successful excavation training samples, which would make the excavation training harder due to more severe class imbalance.
% On the other hand, if we decrease the success threshold to be smaller than $134$ cm$^3$, 
% the excavation classification models would predict excavation samples with bucket filling rates below $30\%$ as successes. This means the learning-based planners would generate excavation trajectories with lower excavation volumes. 
On the other hand, if we decrease the success threshold to be smaller than $134$ cm$^3$, the learning-based planners would be more likely to generate excavation trajectories whose bucket filling rates are below $30\%$. This would hurt the excavation performance of the learning-based planners. 
Therefore, we believe $134$ cm$^3$ is a reasonable success threshold for our excavation learning. 
Moreover, it is shown in Section~\ref{sec:exp_sim} that classification-based CEM-voxel and CEM-RGBD outperform regression-based CEM-voxel-reg and CEM-RGBD-reg respectively, which empirically justify the choice of the excavation threshold.

\subsection{Experimental Excavation Trajectory Analysis}
\begin{figure}[h]
    \centering
    \hfill
    \begin{subfigure}[b]{0.23\textwidth}   
        \centering 
        \includegraphics[width=\textwidth]{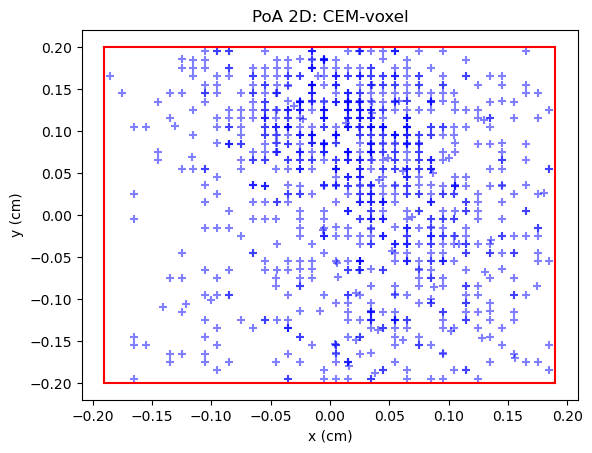}
        \caption{The PoA distribution of CEM-voxel.}  
        %\label{fig:x}
    \end{subfigure}
    \begin{subfigure}[b]{0.23\textwidth}
        \centering
        \includegraphics[width=\textwidth]{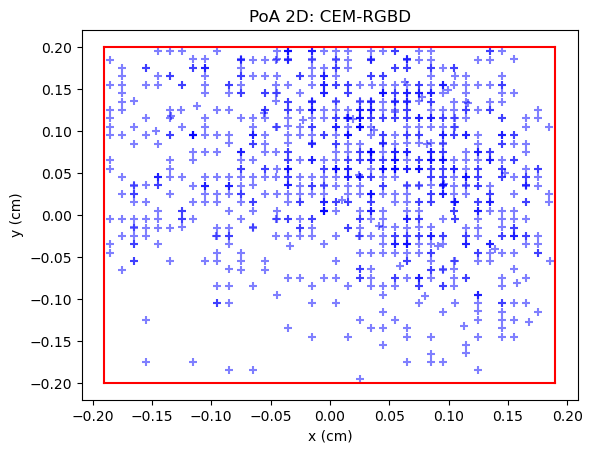}
        \caption{The PoA distribution of CEM-RGBD.}
        %\label{fig:}
    \end{subfigure}
    \vskip\baselineskip
    \hfill
    \begin{subfigure}[b]{0.23\textwidth}   
        \centering 
        \includegraphics[width=\textwidth]{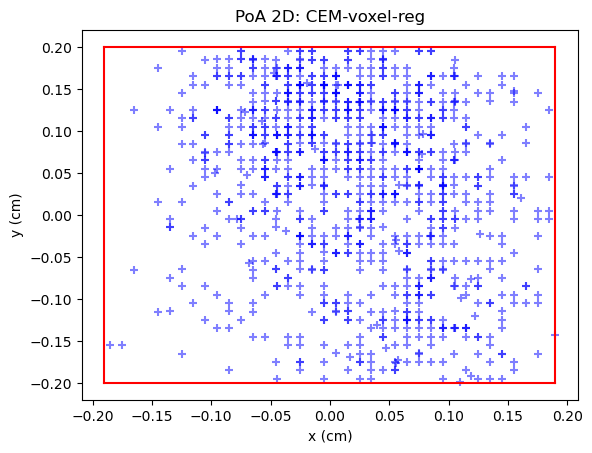}
        \caption{The PoA distribution of CEM-voxel-reg.}  
        %\label{fig:x}
    \end{subfigure}
    \begin{subfigure}[b]{0.23\textwidth}  
        \centering 
        \includegraphics[width=\textwidth]{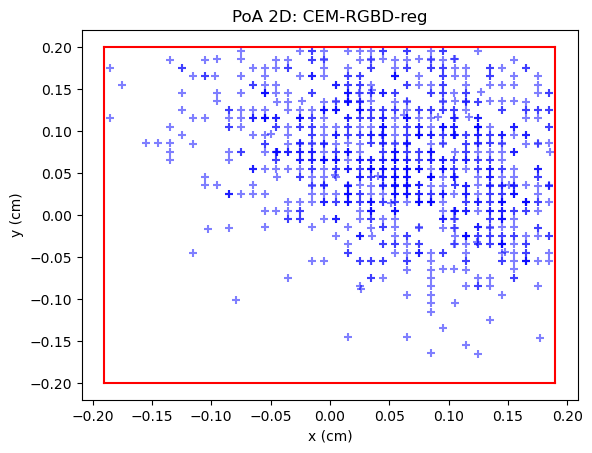}
        \caption{The PoA distribution of CEM-RGBD-reg.} 
        %\label{fig:x}
    \end{subfigure}
    \vskip\baselineskip
    \hfill
    \begin{subfigure}[b]{0.23\textwidth}   
        \centering 
        \includegraphics[width=\textwidth]{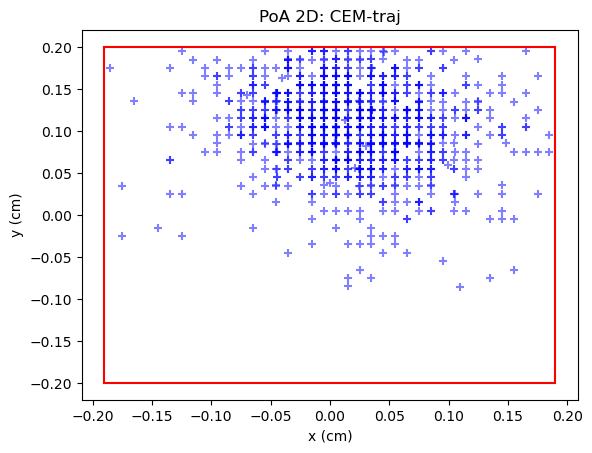}
        \caption{The PoA distribution of CEM-traj.}  
        %\label{fig:x}
    \end{subfigure}
    \begin{subfigure}[b]{0.23\textwidth}   
        \centering 
        \includegraphics[width=\textwidth]{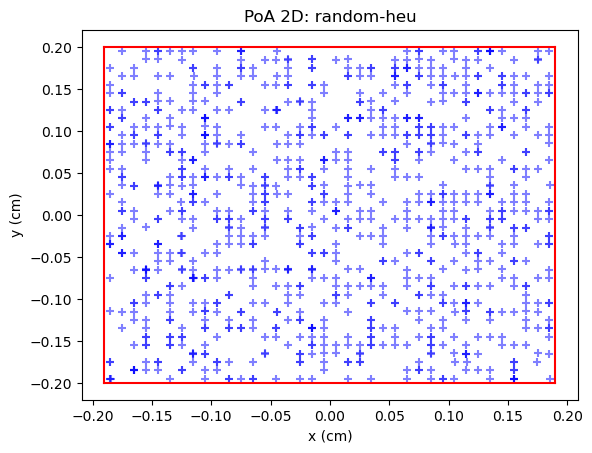}
        \caption{The PoA distribution of random-heu.}  
        %\label{fig:x}
    \end{subfigure}
    %\vskip\baselineskip
    \hfill
    %\centering
    \begin{subfigure}[b]{0.23\textwidth}  
        \centering 
        \includegraphics[width=\textwidth]{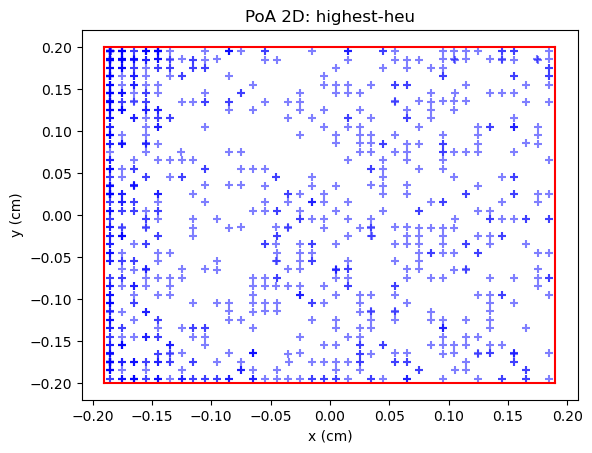}
        \caption{The PoA distribution of highest-heu.} 
        %\label{fig:x}
    \end{subfigure}
    \caption{The 2D PoA distributions of different methods in simulation. The red bounding box represents the 2D tray projected from the overhead view. Each PoA is plotted as a cross in the 2D tray frame.}
    %The origin is at the center of the tray. The robot base locates at $(x=-50, y=0)$ in the 2D tray frame.}
    \label{fig:PoA_sim_200_400}
\end{figure}

The trajectory parameter mean and standard deviation of the $1000$ simulated experimented excavations for each of the seven planners are presented in Table~\ref{table:traj_mean} and~\ref{table:traj_std} respectively. 
In Figure~\ref{fig:PoA_sim_200_400}, we plot the PoA distributions of the simulated experiment results of all seven planners. 
The coordinate origin is at the center of the tray in each PoA plot. The robot base locates at $(x=-50, y=0)$ in the 2D tray frame.

The GTP means and standard deviations of different planners are mostly similar, which shows learning-based planners generate excavation trajectories with large GTP diversity. The PoA standard deviations of the learning-based planners are smaller than heuristic planners due to the randomness of heuristic planners. In terms of the PoA mean, CEM-voxel, CEM-RGBD, and CEM-voxel-reg are similar and they are relatively different from CEM-RGBD-reg, CEM-traj-opt, random-heu, and highest-heu. 

As can be seen from both the trajectory means in Table~\ref{table:traj_mean} and the PoA distribution plots in Figure~\ref{fig:PoA_sim_200_400}, learning-based planners prefer to generate PoA in the top (i.e., positive $y$ direction) right (i.e., positive $x$ direction) area. Highest-heu generates a lot of PoA close to the left edge of the tray (i.e., $x=-0.19$ m). Objects tend to be pushed to the left wall of the tray during excavations. So highest points are more likely to occur close to the left wall of the tray. 

\begin{figure}[h]
    \centering
    \hfill
    \begin{subfigure}[b]{0.23\textwidth}   
        \centering 
        \includegraphics[width=\textwidth]{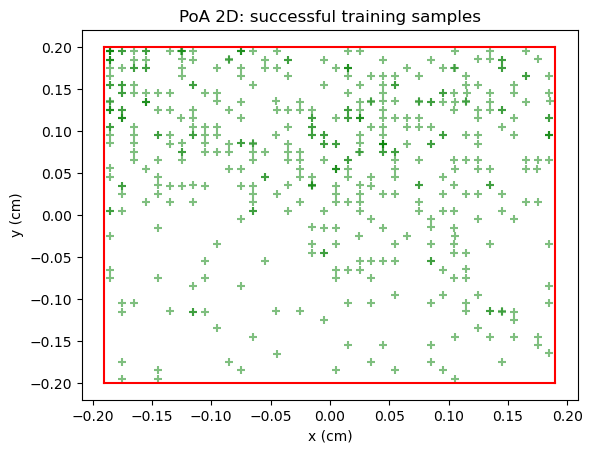}
        \caption{The PoA distribution of successful training samples.}  
        \label{fig:PoA_train_suc}
    \end{subfigure}
    \begin{subfigure}[b]{0.23\textwidth}
        \centering
        \includegraphics[width=\textwidth]{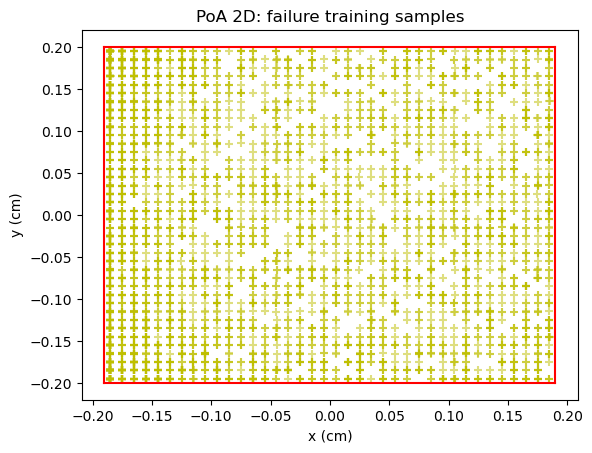}
        \caption{The PoA distribution of failure training samples.}
        \label{fig:PoA_train_fail}
    \end{subfigure}
    \caption{The 2D PoA distributions of the training data in simulation.}
    \label{fig:PoA_train_suc_failure}
\end{figure}

We randomly select $5000$ excavation training excavation samples and plot the PoA of the successful and failure excavations separately in Figure~\ref{fig:PoA_train_suc_failure}. As described in Section~\ref{sec:data_collection}, the training data contains half random-heu and half highest-heu excavations statistically. As shown in Figure~\ref{fig:PoA_train_suc}, there are more successful PoA in the top half of the tray, which explains the learning-based planners prefers PoA in the top area. 
The UR5 excavator swings around the swing center $(x=-0.5$ m, $y=0.109$ m$)$ in the tray 2D coordinate. When the PoA gets closer to the bottom of the tray (i.e., $y=0.2$ m), the robot would have relatively less space to drag and close due to collision with the tray. 
Moreover, there are more failure excavations close to the left edge of the tray in Figure~\ref{fig:PoA_train_fail}, which pushes the learning-based planners to plan PoA away from the left edge.

It has been reported in Section~\ref{sec:exp_sim} that learning-based scene-dependent planners such as CEM-voxel significantly outperform CEM-traj, which shows that it is important to learn to plan scene-dependent excavation trajectories using the visual representation of the excavation scene. However, the trajectory and PoA distributions can not reflect the benefits of the learning-based scene-dependent planning. In the future, we would like to further investigate and understand how learning-based planners use the scene representation to generate high-quality excavation trajectories. 

\begin{table}[h!]
\begin{center}
\resizebox{0.5\textwidth}{!}{
\begin{tabular}{ |c|c|c| } 
 \hline
Method & Trajectory Mean \\ 
 \hline
 CEM-voxel & $[0.03, 0.04, -1.54, 0.12, 0.24, -2.56]$\\
 \hline
 CEM-RGBD & $[0.02, 0.05, -1.59, 0.13, 0.24, -2.79]$ \\
 \hline
 CEM-voxel-reg & $[0.02, 0.04, -1.54, 0.13, 0.25, -2.51]$ \\
 \hline
 CEM-RGBD-reg & $[ 0.06, 0.06, -1.52, 0.14, 0.26, -2.59]$ \\
 \hline
 CEM-traj-opt & $[0.02, 0.1, -1.56, 0.14, 0.23, -2.54]$ \\
 \hline
 random-heu & $[0, 0, -1.57, 0.12, 0.22 -2.61]$ \\
 \hline
 highest-heu & $[-0.06, 0.01, -1.59, 0.12, 0.23, -2.62]$ \\
 \hline
\end{tabular}
}
\caption{The trajectory means of different methods in simulated experiments.}
\label{table:traj_mean}
\end{center}
\end{table}

\begin{table}[h!]
\begin{center}
\resizebox{0.5\textwidth}{!}{
\begin{tabular}{ |c|c|c| } 
 \hline
Method & Trajectory Std \\ 
 \hline
 CEM-voxel & $[0.07, 0.1, 0.34, 0.04, 0.1, 0.36]$\\
 \hline
 CEM-RGBD & $[0.09, 0.08, 0.33, 0.04, 0.11, 0.29]$ \\
 \hline
 CEM-voxel-reg & $[0.07, 0.11, 0.32, 0.04, 0.1, 0.37]$ \\
 \hline
 CEM-RGBD-reg & $[0.07, 0.07, 0.28, 0.04, 0.09, 0.26]$ \\
 \hline
 CEM-traj-opt & $[0.06, 0.05, 0.36, 0.03, 0.1 0.37]$ \\
 \hline
 random-heu & $[0.11, 0.11, 0.3, 0.04, 0.1, 0.31]$ \\
 \hline
 highest-heu & $[0.12, 0.13, 0.3, 0.04, 0.1, 0.3]$ \\
 \hline
\end{tabular}
}
\caption{The trajectory standard deviations of different methods in simulated experiments.}
\label{table:traj_std}
\end{center}
\end{table}

% $ rosrun tf tf_echo ur5_base_link bucket_tooth
% At time 1621387410.509
% - Translation: [0.720, 0.109, 0.170]
% - Rotation: in Quaternion [0.612, -0.612, 0.354, -0.355]
%             in RPY (radian) [-2.091, 0.000, -1.570]
%             in RPY (degree) [-119.794, 0.023, -89.960]

\subsection{Ablation Experiments}
\label{sec:ablation}

% I am wondering whether the authors can add some analysis or ablation study to show insights on how the learning-based planners excel, e.g., showing a distribution of (x,
% y) generated by the learning-based planners and the heuristic planners,
% or showing the results of the heuristic planners by using random (x, y)
% but exact same ($\alpha$, d, l, $\beta$) generated by the learning-based
% planners.

% Given the $(x, y)$ coordinate of the attacking pose $p$, its $z$ coordinate value on the objects clutter surface is computed as the height of the gridmap of the objects in clutter at $(x, y)$. More details and examples of the grid map can be found in Section~\ref{sec:data_collection}. We fix the lifting height $h$ as the robot base height. Therefore, we can represent a task space excavation trajectory $T$ using $6$ parameters $T=(x, y, \alpha, d, l, \beta)$. 

Ablation experiments are performed to show insights on how the learning-based planners improve excavation for cluttered rigid objects. 
% As defined in Section~\ref{sec:task_traj}, a task space excavation trajectory $T$ contains $6$ parameters $T=(x, y, \alpha, d, l, \beta)$. We name $(x, y)$ and $(\alpha, d, l, \beta)$ of a task trajectory $T$ as ``Point of Attack (PoA)" and ``Geometric Trajectory Parameters (GTP)" respectively for the ablation study.
The ablation study is focused on the CEM-voxel planner, since it achieves the best excavation performance in the simulated and real-robot experiments.
% We perform two ablation experiments for CEM-voxel are performed. 
Two ablation experiments are performed by replacing the PoA and GTP of each CEM-voxel trajectory with random parameters respectively. 
Random PoA and GTP parameters are uniformly sampled from the same range as the heuristic planners introduced in Section~\ref{sec:heu_planner}.
$1000$ excavation trials are experimented for both ablation experiments in simulation using the same experiment setup and protocol as Section~\ref{sec:exp_setup_sim}.
% $100$ excavation episodes are experimented for each method. $10$ excavation trials are sequentially performed for each excavation episode. That gives us $1000$ excavation experimented trials in total for each method. 

% the excavation volume, the excavated objects number, and the excavation success rate. Same as model training in Section~\ref{sec:exv_train}, if the total volume of a sample's successfully excavated objects is above $134 cm^{3}$, we count the excavation sample as a success, otherwise a failure. We also report the computation time of each planner. 

% We list the mean with standard deviation in parentheses for all metrics except the success rate.

The excavation volumes, the excavated objects numbers, and the excavation success rates of both ablation experiments are presented in Table~\ref{table:exv_exp_sim_ablation}. We list the mean with standard deviation in parentheses for excavation volumes and objects numbers.
The original CEM-voxel experiment results in simulation are shown in Table~\ref{table:exv_exp_sim_200_400}. 
CEM-voxel with random PoA and random GTP both performs worse than the original CEM-voxel in terms the three excavation metrics. This demonstrates CEM-voxel learns about how to generate both good PoA and GTP parameters for excavation.
CEM-voxel with random PoA gets worse excavation performance than CEM-voxel with random GTP. This implies the learning of PoA matters more than the learning of GTP for CEM-voxel. 

\begin{table}[h!]
\begin{center}
\resizebox{0.5\textwidth}{!}{
\begin{tabular}{ |c|c|c|c|c| } 
 \hline
Method & Volume (cm$^3$) & Number & Success Rate \\ 
 \hline
 CEM-voxel with random PoA & $104.56$ $(99.91)$ & $5.8$ $(5.72)$ & $36.2\%$ \\
 \hline
 CEM-voxel with random GTP  & $119.62$ $(92.79)$ & $6.77$ $(5.34)$ & $42.7\%$ \\
 \hline
%  CEM-traj-opt & $97.27$ $(100.73)$ & $5.54$ $(5.87)$ & $32.4\%$ \\
%  \hline
\end{tabular}
}
\caption{Ablation experiment results in simulation.}
\label{table:exv_exp_sim_ablation}
\end{center}
\end{table}

\label{sec:appendix}
\end{document}